%% file: main.tex
\pdfoutput=1

\documentclass[11pt]{article}
\usepackage{styles/acl}
\usepackage{times}
\usepackage{latexsym}
\usepackage[T1]{fontenc}
\usepackage[utf8]{inputenc}
\usepackage{microtype}
\usepackage{inconsolata}
\usepackage{graphicx}
\usepackage{amsmath} 
\usepackage{amsfonts}
\usepackage{amssymb}
\usepackage{algorithm}
\usepackage{algpseudocode}
\usepackage{booktabs}
\usepackage{color,colortbl}
\usepackage{xcolor}
\usepackage{makecell}
\usepackage{tcolorbox}
\definecolor{COLOR_MEAN}{HTML}{f0f0f0}
\usepackage{multirow}
\usepackage{etoc}
\usepackage{hyperref}
\usepackage[misc]{ifsym}
\definecolor{titlepink}{RGB}{255,105,110}

\title{MoLoRAG: Bootstrapping Document Understanding \\ via Multi-modal Logic-aware Retrieval}

\author{
    Xixi Wu$^{1}$, Yanchao Tan$^{2}$, Nan Hou$^{1}$, Ruiyang Zhang$^{3}$, Hong Cheng$^{1}$$^{(\textrm{\Letter})}$ \\
    $^1$The Chinese University of Hong Kong \\
    $^2$Fuzhou University \quad
    $^3$University of Macau \\ 
    {\texttt{\{xxwu, nhou, hcheng\}@se.cuhk.edu.hk}} \\
    \texttt{yctan@fzu.edu.cn, yc47931@um.edu.mo}
}

\begin{document}
\maketitle
\begin{abstract}

Document Understanding is a foundational AI capability with broad applications, and Document Question Answering (DocQA) is a key evaluation task. Traditional methods convert the document into text for processing by Large Language Models (LLMs), but this process strips away critical multi-modal information like figures. While Large Vision-Language Models (LVLMs) address this limitation, their constrained input size makes multi-page document comprehension infeasible. Retrieval-augmented generation (RAG) methods mitigate this by selecting relevant pages, but they rely solely on semantic relevance, ignoring logical connections between pages and the query, which is essential for reasoning.

To this end, we propose \textbf{MoLoRAG}, a \underline{lo}gic-aware retrieval framework for \underline{m}ulti-m\underline{o}dal, multi-page document understanding. By constructing a page graph that captures contextual relationships between pages, a lightweight VLM performs graph traversal to retrieve relevant pages, including those with logical connections often overlooked. This approach combines semantic and logical relevance to deliver more accurate retrieval. After retrieval, the top-$K$ pages are fed into arbitrary LVLMs for question answering. To enhance flexibility, MoLoRAG offers two variants: a training-free solution for easy deployment and a fine-tuned version to improve logical relevance checking. Experiments on four DocQA datasets demonstrate average improvements of \textbf{9.68\%} in accuracy over LVLM direct inference and \textbf{7.44\%} in retrieval precision over baselines. Codes and datasets are released at \href{https://github.com/WxxShirley/MoLoRAG}{\texttt{\small{https://github.com/WxxShirley/MoLoRAG}}}.
\end{abstract}

\input{sections/1-intro} 
\input{sections/2-related_works}
\input{sections/3-method}
\input{sections/4-experiment}

\section{Conclusion}

In this paper, we tackle the DocQA task by addressing the limitations of prior methods that rely only on semantic relevance for retrieval. By incorporating logical relevance, our VLM-powered retrieval engine performs multi-hop reasoning over page graph to identify key pages. Extensive experiments demonstrate that MoLoRAG delivers superior retrieval accuracy, achieves SOTA performance, and ensures seamless compatibility with LVLMs.

\section*{Acknowledgments}
This research is supported in part by project \#MMT-p2-23 of the Shun Hing Institute of Advanced Engineering, The Chinese University of Hong Kong, by grants from the Research Grants Council of the Hong Kong SAR, China (No. CUHK 14217622). This research is also supported in part by the National Natural Science Foundation of China (No.62302098), and the Fujian Provincial Natural Science Foundation of China (2025J01540). The authors would like to express their gratitude to the reviewers for their valuable feedback, which has improved the clarity and contribution of the paper. 

\section*{Limitations}
MoLoRAG primarily focuses on closed-domain document understanding, where the relevant document is provided. Extending this approach to an \textbf{open-domain} setting, where the document corpus consists of extensive and diverse documents, is a challenge. This is because modeling relationships not only within individual document but also across different documents, as well as performing graph traversal between both document- and page-level nodes, becomes complex. 

\noindent In this paper, we did not use any non-public data, unauthorized software, or APIs, and there are no privacy or other related ethical concerns associated with our work. 

\bibliography{reference}

\clearpage 
\newpage 

\appendix

\input{appendix/1-pseudo-code}
\input{appendix/2-prompt}
\input{appendix/3-train_star}
\input{appendix/4-supple_experiment}

\end{document}

%% file: sections/1-intro.tex
\section{Introduction}

\begin{figure}[!t]
    \centering
    \includegraphics[width=\linewidth]{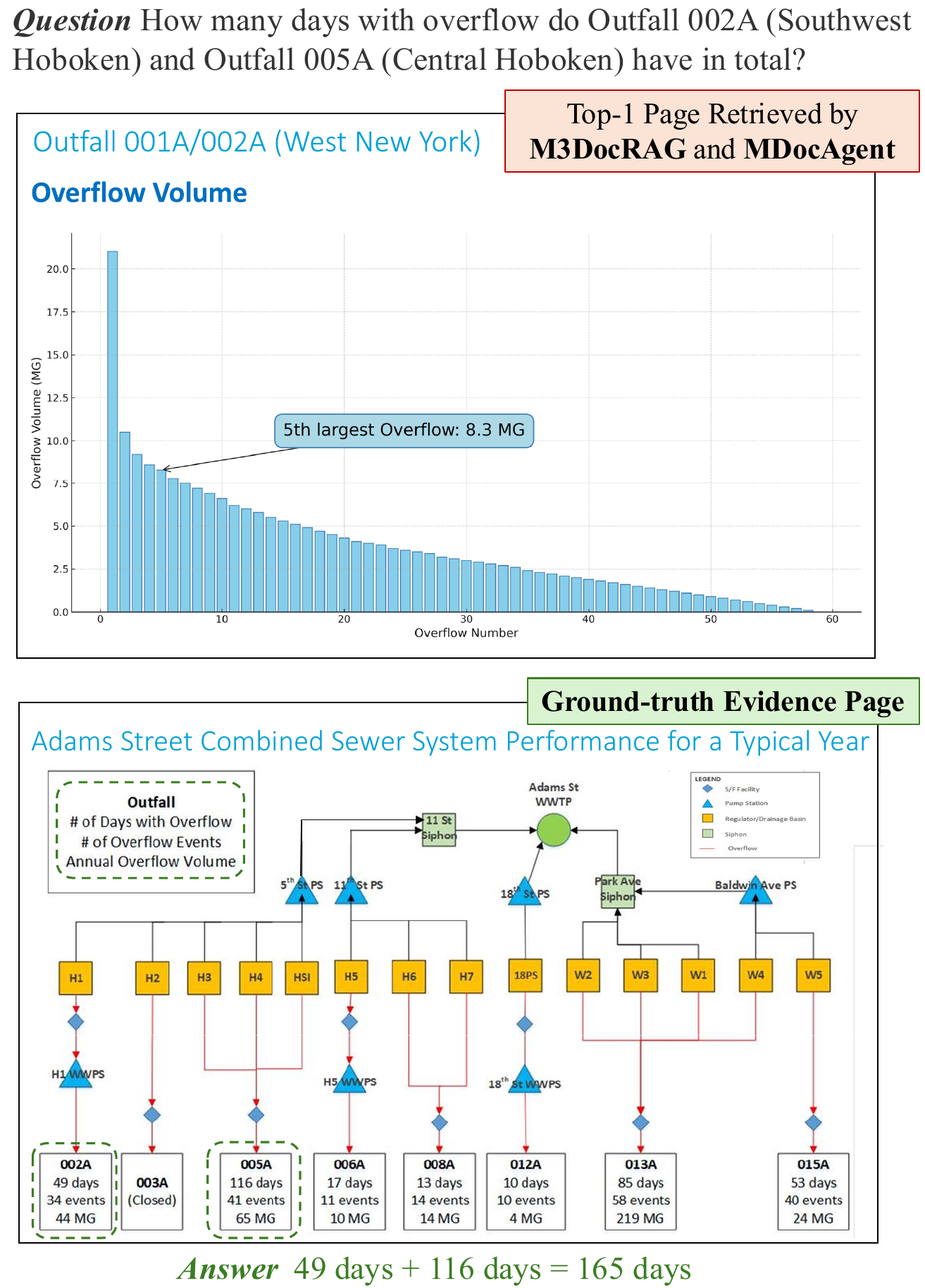}
    \vspace*{-12pt}
    \caption{\textbf{Illustration of a retrieval example on LongDocURL \cite{deng2024longdocurl}.} Both M3DocRAG \cite{cho2024m3docrag} and MDocAgent \cite{han2025mdocagent} rely solely on \textbf{semantic relevance} between the query and the page for retrieval. As a result, they retrieve a page containing keywords from the question but lacking the necessary information to answer it. In contrast, the ground-truth evidence page, successfully retrieved by our MoLoRAG, is \textbf{logically relevant} to the question, providing detailed statistics for each outfall and enabling accurate derivation of the correct answer. 
    }
    \label{fig:toy_example}
    \vspace*{-12pt}
\end{figure}

Document Understanding is a foundational AI capability with extensive real-world applications, such as interpreting medical reports and assisting with academic literature. This ability holds significant potential to improve productivity and support decision-making ~\cite{ding2022vdoc, ma2024visa, suri2024visdom, zhang2024ocr}. A key task for evaluating document understanding is Document Question Answering (DocQA), which requires models to automatically answer questions based on the content of a document. 

Classic approaches to DocQA typically follow a two-step pipeline: the document is first converted into text using Optical Character Recognition (OCR) \cite{review_ocr, MinuerU, GOT-OCR}, and then Retrieval-augmented Generation (RAG) techniques identify relevant paragraphs to feed into Large Language Models (LLMs) for question answering. However, the text extraction process often strips away essential multi-modal information, such as tables, figures, and document layouts, resulting in incomplete document understanding. Large Vision-Language Models (LVLMs) address this limitation by processing image-format document snapshots, enabling multi-modal comprehension. Nevertheless, LVLMs, such as LLaMA-Vision ~\cite{grattafiori2024llama} and LLaVA-Next~\cite{li2024llavanext-strong}, are constrained to single-image inputs, rendering them ineffective for long, multi-page documents.

Recent research has explored methods to address these challenges. For example, M3DocRAG ~\cite{cho2024m3docrag} leverages a document encoder, i.e., ColPali ~\cite{faysse2024colpali}, to encode individual pages and retrieve relevant ones based on vector similarity. This approach reduces the number of input pages, alleviating the comprehension burden for LVLMs. MDocAgent ~\cite{han2025mdocagent} extends this by introducing parallel pipelines for text and image retrieval, with specialized agents for each modality to enable collaborative reasoning. While effective, these methods focus primarily on \textbf{semantic relevance}, matching queries to pages based on embedding similarity. For example, as shown in Figure \ref{fig:toy_example}, when asked to determine the total overflow days for two outfalls, the top-$1$ page retrieved by both methods contains keywords from the question but lacks detailed information about each outfall. In contrast, a \textbf{logically relevant} page, such as the ground-truth evidence page, provides detailed statistics for each outfall, enabling reasoning (e.g., summing the overflow days) to derive the correct answer.

In DocQA, accurate retrieval is critical, as it directly impacts downstream answering.  Without precise retrieval, LVLMs are prone to errors or hallucinations stemming from irrelevant or incomplete inputs. Addressing this challenge requires retrieval methods that go beyond surface-level semantic matching to capture deeper logical relationships. Building on this insight, we propose \textbf{MoLoRAG}, a graph-based retrieval framework tailored for \underline{\textbf{M}}ulti-m\underline{\textbf{o}}dal \underline{\textbf{Lo}}gic-aware document understanding. Document pages naturally exhibit structured relationships, e.g., cross-references and shared entities. Leveraging this property, we first construct a page graph to represent the dependencies between pages. A lightweight VLM serves as the retrieval engine, reasoning over this graph through traversal to identify logically relevant pages. Finally, both semantic and logical relevance are combined into a unified similarity score to re-rank pages, enabling a more comprehensive retrieval process. 

To further enhance its utility, MoLoRAG introduces two variants to offer flexibility for different deployments. The first is a training-free variant, which leverages a pre-trained VLM, e.g., Qwen2.5-VL-3B~\cite{Qwen2.5-VL}, to perform graph traversal and retrieval directly, providing an off-the-shelf solution that is easy to deploy. The second is a fine-tuned variant, which involves training the retrieval engine on a curated dataset to improve its reasoning capabilities. This fine-tuned version functions as a more intelligent retrieval engine, capable of capturing nuanced relationships between queries and document pages. Moreover, MoLoRAG demonstrates strong compatibility with arbitrary LVLMs. Once the retrieval step is complete, only the top-$K$ page snapshots are passed to an LVLM for question answering, filtering out irrelevant content and ensuring concise, high-quality inputs. To summarize, our contributions are as follows: 

\begin{itemize}
    \item \textbf{Logic-aware Retrieval Framework} We highlight the importance of page retrieval in DocQA and propose MoLoRAG, a novel retrieval method that incorporates logical relevance. By representing the document as a page graph and enabling a VLM to perform multi-hop reasoning through graph traversal, our method identify both semantically and logically relevant pages. 

    \item \textbf{Comprehensive Experiments} We conduct extensive experiments on four DocQA datasets, comparing MoLoRAG with LLM-based, LVLM-based, and Multi-agent methods. Results demonstrates its superior retrieval accuracy, significant performance improvements over baselines, and flexible compatibility with arbitrary LVLMs.

    \item \textbf{Released Model and Dataset} We release the fine-tuned retriever engine model weights\footnote{\href{https://huggingface.co/xxwu/MoLoRAG-QwenVL-3B}{https://huggingface.co/xxwu/MoLoRAG-QwenVL-3B}} and the curated training dataset\footnote{\href{https://huggingface.co/datasets/xxwu/MoLoRAG}{https://huggingface.co/datasets/xxwu/MoLoRAG}}, empowering further development of intelligent and logic-aware retrieval engines. 
\end{itemize}

%% file: sections/2-related_works.tex
\section{Related Works}

\textbf{Document Question Answering} DocQA is a core task for evaluating document understanding. Early benchmarks \cite{mathew2021docvqa, tito2023mpdocvqa, SlideVQA2023} focused on single-page or short documents with low information density, where questions targeted individual elements like text or figures. Recent benchmarks \cite{ma2024mmlongbench, deng2024longdocurl} shift toward lengthy, information-rich documents, introducing challenges like cross-page and multi-modal reasoning. DocQA methods can be broadly categorized into two branches based on backbone models: LLM-based and LVLM-based. LLM-based methods rely on OCR techniques to extract text from the document, enabling text-based question answering. LVLM-based methods, on the other hand, leverage their inherent multi-modal capabilities to process document images directly. With the advancements in LVLMs, the latter approach now dominates recent solutions. A notable advancement is MDocAgent \cite{han2025mdocagent}, which represents a new category by combining both LLMs and LVLMs into a multi-agent framework for collaborative question answering. However, challenges like input size limitations still necessitate effective retrieval strategies to reduce input burden and enhance performance. 

\noindent \textbf{Retrieval-augmented Generation} RAG enhances LLMs by supplementing them with external knowledge, improving performance in domain-specific or knowledge-intensive tasks~\cite{gao2024ragsurvey, lewis2021retrieval, asai2024selfrag}. The emergence of LVLMs has further expanded RAG to multi-modal contexts, enabling the retrieval of relevant images to handle knowledge-seeking queries \cite{chen2024mllm, chen2022murag}. Despite these advancements, existing RAG methods fail to address the unique challenges of DocQA, involving highly interleaved textual and visual elements. For page retrieval in DocQA, existing method like M3DocRAG~\cite{cho2024m3docrag} rely on document encoders for semantic-based retrieval, neglecting the logical relevance essential for accurate question answering.

\noindent \textbf{Graph-based RAG} GraphRAG is an advanced RAG paradigm that leverages graph-structured knowledge and retrieval for improved contextual reasoning \cite{zhang2025survey,xiang2025use}. Existing methods are categorized into two types: Knowledge-based, which constructs knowledge graphs through entity recognition and relation extraction \cite{he2024gretriever}, and Index-based, which creates a two-layer graph linking high-level topic nodes to detailed text nodes for efficient retrieval~\cite{sarthi2024raptor, edge2025graphrag, liu2025hoprag, li-etal-2024-graphreader}. However, current GraphRAG approaches are limited to text and cannot handle the document with multi-modal information. We are the first to extend GraphRAG to the document domain by constructing a page graph that enables reasoning over its structure.

%% file: sections/3-method.tex
\begin{figure*}[!t]
    \centering
    \includegraphics[width=0.95\linewidth]{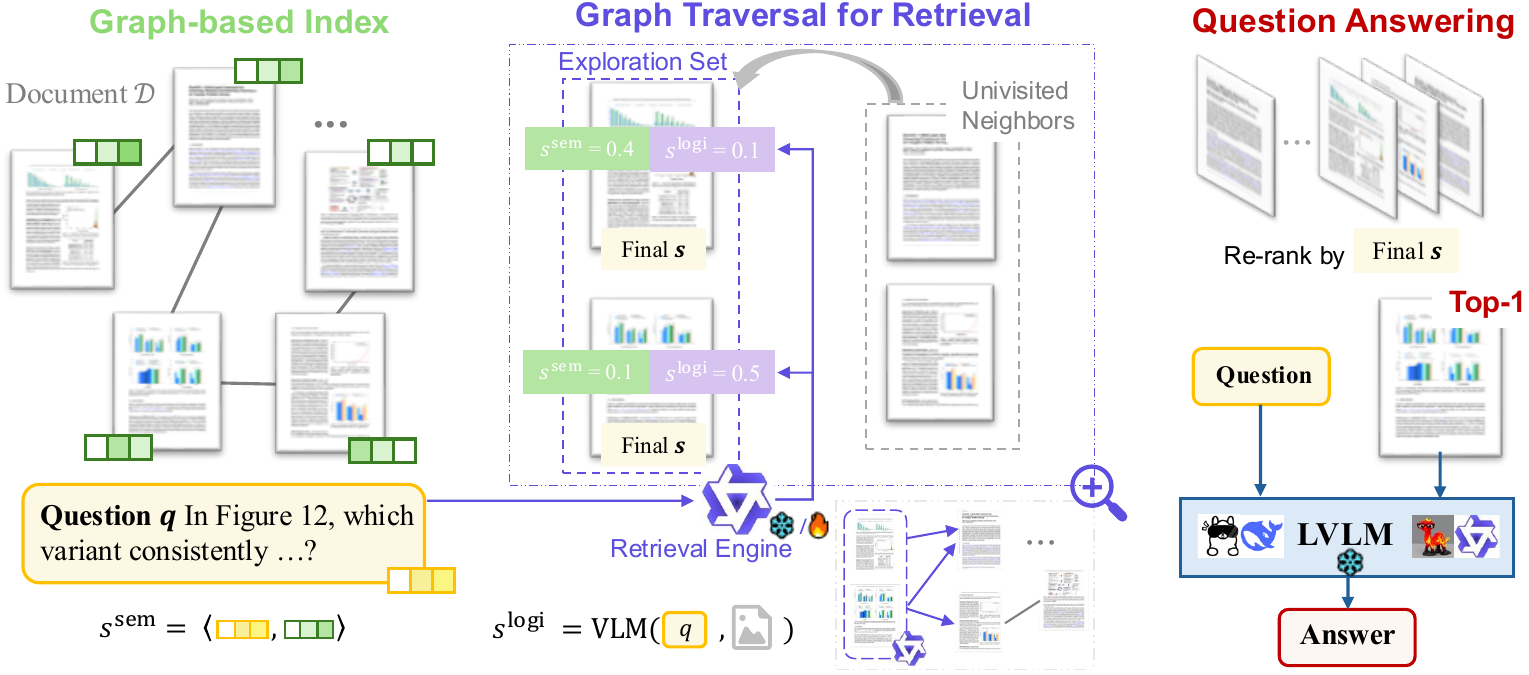}
    \caption{\textbf{Illustration of MoLoRAG framework.}}
    \label{fig:method}
    \vspace*{-8pt}
\end{figure*}

\section{Methodology}

In this section, we present the details of MoLoRAG, a novel graph-based retrieval framework designed to facilitate multi-modal and multi-page document understanding. The overall framework is illustrated in Figure~\ref{fig:method}. 

\subsection{Preliminary}
Given a question $q$ expressed in natural language and a document $\mathcal{D} = \{ p_1, p_2, \ldots, p_N  \}$, where each $p_i$ represents an individual page in the form of an RGB image, and $N$ is the total number of pages. The goal of DocQA is to generate an answer $a$ that accurately addresses $q$ using the information contained within $\mathcal{D}$. To solve this task, MoLoRAG adopts an LVLM-based two-stage framework: 
\begin{itemize}
    \item \textbf{Retrieval:} Given the extensive nature of document $\mathcal{D}$, the first step involves retrieving the top-$K$ most relevant pages for the question, denoted as $\mathcal{P}^{r} = \{ p_1^{r}, \ldots , p_K^r\}$, where $K\ll N$ (e.g., $K=3$). Unlike traditional retrieval methods that rely solely on semantic relevance, MoLoRAG incorporates both semantic and logical relevance to enhance retrieval accuracy and contextual understanding for effective reasoning.

    \item \textbf{Generation:} The retrieved pages $\mathcal{P}^r$, along with the input question $q$, are then fed into an LVLM to generate the answer $a$. For LVLMs that cannot directly process multiple images, we use a processing function $\text{Process}(\cdot)$ to prepare $\mathcal{P}^r$, e.g., concatenating multiple images into a single composite one. Formally, this stage is expressed as:

    \begin{equation*}
        a = \text{LVLM}(q, \text{Process}(\mathcal{P}^r)).
    \end{equation*}

\end{itemize}

\subsection{Logic-aware Page Retrieval} 
In this subsection, we detail the retrieval process of MoLoRAG. We first construct a page graph as a graph-based index, depicting the relationships between pages within a document. Then, a VLM serves as the retrieval engine, performing reasoning over this graph through traversal to adaptively identify pages that are both semantically and logically relevant to the given question.

\noindent \textbf{Graph-based Index} Firstly, each document page $p_i$ is encoded into a latent embedding that captures its distinct multi-modal content, represented as $E_{p_i} = \text{DocEncoder}(p_i) \in \mathbb{R}^{k \times d}$, where $k$ denotes the number of visual tokens per page and $d$ is the embedding dimension. Following \citet{cho2024m3docrag, han2025mdocagent}, we choose ColPali~\cite{faysse2024colpali} as the document encoder due to its demonstrated effectiveness in preserving multi-modal semantics. 

Using these embeddings, we construct a page graph $G(\mathcal{V}, \mathcal{E})$ to represent relationships between pages. In this graph, each node $p_i \in \mathcal{V}$ corresponds to a page from the document $\mathcal{D}$, and edges $\mathcal{E}$ are established between pairs of pages based on their similarity. Specifically, an edge $(p_i, p_j)$ is added if the similarity between their embeddings exceeds a threshold $\theta$, expressed as: $\mathcal{E}= \{ (p_i, p_j) | \langle E_{p_i}, E_{p_j} \rangle \geq \theta\}$ where $\langle \cdot, \cdot \rangle$ denotes the inner product as the similarity measure. While such graph construction mechanism is simple, it offers the advantages of being \textbf{efficient}, \textbf{automatic}, ensuring \textbf{scalability} to large document, and leveraging prior knowledge encoded in the embedding.

\noindent \textbf{Graph Traversal for Retrieval} With the page graph constructed, we leverage a VLM as the retrieval engine to evaluate the relevance of each visited page in relation to the given question. This approach overcomes the limitations of traditional semantic-only retrieval by incorporating logical checking into the process. By utilizing the reasoning capabilities of the VLM, our method effectively identifies important pages that may otherwise be overlooked. The graph traversal process is outlined as follows, aligning with the pseudo-code in Algorithm \ref{algorithm:beamsearch} in the Appendix.

\noindent \textbf{$-$ Initialization} For the question $q$ and a page $p_i$ from the document, the document encoder computes a semantic relevance score as $s_i^{\text{sem}} = \langle \text{DocEncoder}(q), E_{p_i} \rangle$. Based on these scores, the top-$w$ nodes (pages) with the highest semantic scores are selected as the initial exploration set. 

\noindent \textbf{$-$ Relevance Scoring} For page $p_i$ in the exploration set, the VLM assigns a logical relevance score $s_i^{\text{logi}}$ using the prompt provided in Appendix \ref{appendix:prompt}, reflecting the deeper logical connection of the page to the question. The final relevance score $s_i$ is then updated as $s_i = \text{Combine}(s^{\text{sem}}_i, s^{\text{logi}}_i)$, where $\text{Combine}(\cdot)$ integrates semantic and logical relevance scores, e.g., taking their weighted average.

\noindent \textbf{$-$ Iterative Traversal} The traversal proceeds iteratively: at each step, we define the candidate set as the unvisited neighbors of the current exploration set. Each page in this candidate set is evaluated and its relevance score is updated using the same combination of semantic and logical relevance. The pages are ranked by their final relevance scores, and only the top-$w$ nodes are retained as the new exploration set for the next iteration. The traversal continues until either the candidate set is empty or the maximum hop limit is reached. Both the exploration set size $w$ and the hop limit $n_{\text{hop}}$ constrain the traversal space, ensuring efficiency by avoiding the exhaustive process of sequentially traversing every page in the document. 

Once the traversal is complete, all visited nodes are \textbf{re-ranked} based on their final relevance scores, and the top-$K$ pages are selected for the subsequent question-answering phase.

\begin{figure}[!t]
    \centering
    \includegraphics[width=\linewidth]{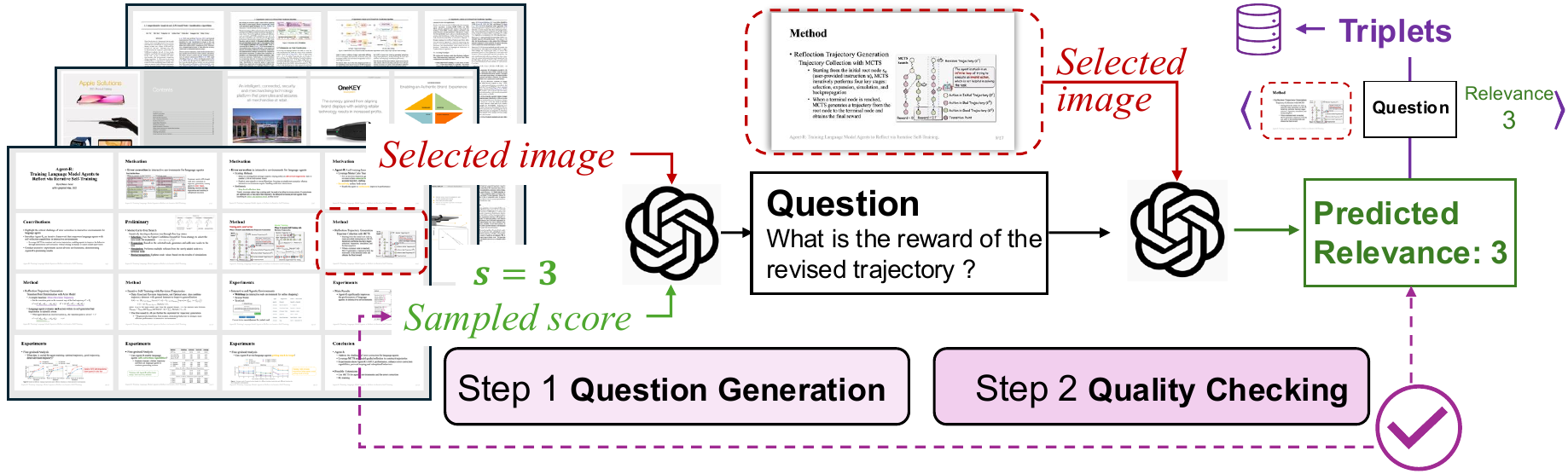}
    \vspace*{-10pt}
    \caption{\textbf{Illustration of training data generation for MoLoRAG+.}}
    \label{fig:data_pipeline}
    \vspace*{-10pt}
\end{figure}

\subsection{Training-required Variant}
While the MoLoRAG framework allows the use of a pre-trained VLM as an off-the-shelf solution for fast deployment, we propose an enhanced variant, \textbf{MoLoRAG+}\footnote{We use \textbf{MoLoRAG+} to denote the fine-tuned version.}. This variant fine-tunes the VLM (retrieval engine) to bolster its reasoning capabilities during graph traversal, enabling the model to assign more accurate logical relevance scores.

\noindent \textbf{Data Preparation (Figure \ref{fig:data_pipeline})} The success of fine-tuning relies on the availability of high-quality training data \cite{sun2024dialinsight}. To achieve this, we utilize GPT-4o \cite{openai2024gpt4ocard} as a data generation engine to create reliable triplets in the format $\langle \texttt{Question}, \texttt{Image}, \texttt{Relevance\_Score} \rangle$, where the \texttt{Relevance\_Score} quantifies the alignment between the question and the image content. These triplets serve as supervision signals for fine-tuning, enabling the model to better estimate logical relevance.  The data creation process begins by randomly selecting a page snapshot (image) from the document and sampling a relevance score from a pre-defined range. Using both the selected image and the sampled relevance score as context, GPT-4o generates a question that reflects the degree to which the selected image can answer it. GPT-4o then predicts the relevance score between the generated question and the image, enabling an automated \textbf{quality-checking}: only samples where the predicted score and the target score closely match (e.g., within a tolerance of $\leq 1$) are retained. To further ensure accuracy, these filtered samples undergo manual verification, ensuring that the final dataset fully aligns with the task’s requirements. Note that the data engine can be replaced with any arbitrary LVLMs instead of proprietary models like GPT-4o to reduce costs. An additional analysis is provided in Appendix \ref{appendix:fine_tune_mmgraph}.

\noindent \textbf{Model Training} Using the curated dataset, we fine-tune the backbone VLM with supervised fine-tuning (SFT) techniques \cite{hu2022lora}. Detailed training configurations are provided in Appendix \ref{appendix:fine_tune_mmgraph}. After fine-tuning, the updated VLM replaces the original pre-trained model as the retrieval engine. By incorporating enhanced logical checking capabilities, this variant is expected to deliver more accurate retrieval performance.

\subsection{Summary}

In the proposed MoLoRAG framework, the top-$K$ scored pages are fed into an LVLM during the question-answering phase, ensuring that only the most relevant information is utilized. Its key strengths include compatibility with arbitrary LVLMs, making it particularly adaptable for models limited to processing a single image by transforming an otherwise infeasible task into a practical solution. By incorporating both semantic and logical relevance, the framework enhances retrieval accuracy (Section \ref{sec:retrieve_comp}). Furthermore, the graph-based traversal mechanism effectively narrows the search space, prioritizing relevant pages and significantly accelerating the retrieval process compared to exhaustive page-by-page traversal (Appendix \ref{appendix:efficiency}). Collectively, these features position MoLoRAG as a powerful solution for the DocQA task.

%% file: sections/4-experiment.tex
\section{Experiments}

\subsection{Experimental Setup}
\noindent \textbf{Datasets} We utilize four datasets from three benchmarks for evaluation, including \textbf{MMLongBench}~\cite{ma2024mmlongbench}, \textbf{LongDocURL}~\cite{deng2024longdocurl}, and \textbf{PaperTab} and \textbf{FetaTab} from the UDA-Benchmark~\cite{hui2024uda}. Dataset statistics are shown in Table~\ref{tab:dataset}. These datasets span a wide range of topics (e.g., administrative files, tutorials, research reports) and feature diverse multi-modal elements (e.g., chart, text, and table). Additionally, they vary in average document length and information density, ensuring a comprehensive evaluation. Other benchmarks like DocVQA \cite{mathew2021docvqa, SlideVQA2023, masry-etal-2022-chartqa} are omitted due to their shorter document lengths and lower information density.

\begin{table}[!t]
    \centering
    \caption{\textbf{Statistics of experimental datasets.}} 
    \vspace*{-8pt}
    \resizebox{\linewidth}{!}{
      \begin{tabular}{c|c|c|c|c}
        \toprule 
       \rowcolor{COLOR_MEAN} \textbf{Dataset} & \# \textbf{Question}  & \# \textbf{Document} & \textbf{Avg. Pages} & \textbf{Avg. Tokens}   \\ \midrule 
        \textbf{PaperTab} & 393 & 307 & 11.0 & 12,685.4 \\ 
        \textbf{FetaTab} & 1,016 & 871 & 15.8 & 16.524.5 \\
        \textbf{MMLongBench} & 1,082 & 135 & 47.5 & 24,992.6  \\
         \textbf{LongDocURL} & 2,325 & 396 & 85.6 & 56,715.1 \\ 
        \bottomrule
    \end{tabular}   
    }
    \label{tab:dataset}
\end{table}

\noindent \textbf{Evaluation Metrics} For MMLongBench and LongDocURL, we follow their original evaluation protocol, using a generalized \textbf{Accuracy} with rule-based evaluation to handle various answer types. Additionally, we report \textbf{Exact Match (EM)} as a supplementary metric, as the answers in these datasets are typically short and concise. For PaperTab and FetaTab, where ground-truth answers are formulated as long sentences or multiple choices, we follow MDocAgent to employ GPT-4o as the evaluator. Specifically, it evaluates \textbf{Binary Correctness} by determining whether the generated answer matches the ground-truth answer, assigning a binary score of 0 or 1. We also evaluate retrieval-stage accuracy using metrics like Recall@$K$, with further details provided in Appendix \ref{appendix:metrics}.
\input{tables/main_top3}

\noindent \textbf{Baselines} We consider the following baselines: (1) \textbf{LLM w. Text RAG} first converts the document into texts using OCR and then applies retrieval techniques to the text, with LLMs serving as the backbone for question answering. (2) \textbf{LVLM Direct Inference} directly feeds LVLMs with full document snapshot images for question answering. For LVLMs that only support single-image input, we follow \citet{ma2024mmlongbench} by concatenating all images into a single combined one. (3) \textbf{M3DocRAG} \cite{cho2024m3docrag} uses ColPali as a page retriever to identify relevant pages and feeds only the retrieved pages to the LVLM for further processing. (4) \textbf{MDocAgent} \cite{han2025mdocagent} is a strong baseline for document understanding that employs a multi-agent system. A text agent and an image agent independently handle their respective modalities and collaborate to synthesize the final answer. Due to space limits, implementation details of each method are provided in Appendix \ref{appendix:implementation}.

\begin{figure*}[!t]
    \centering
    \includegraphics[width=0.95\linewidth]{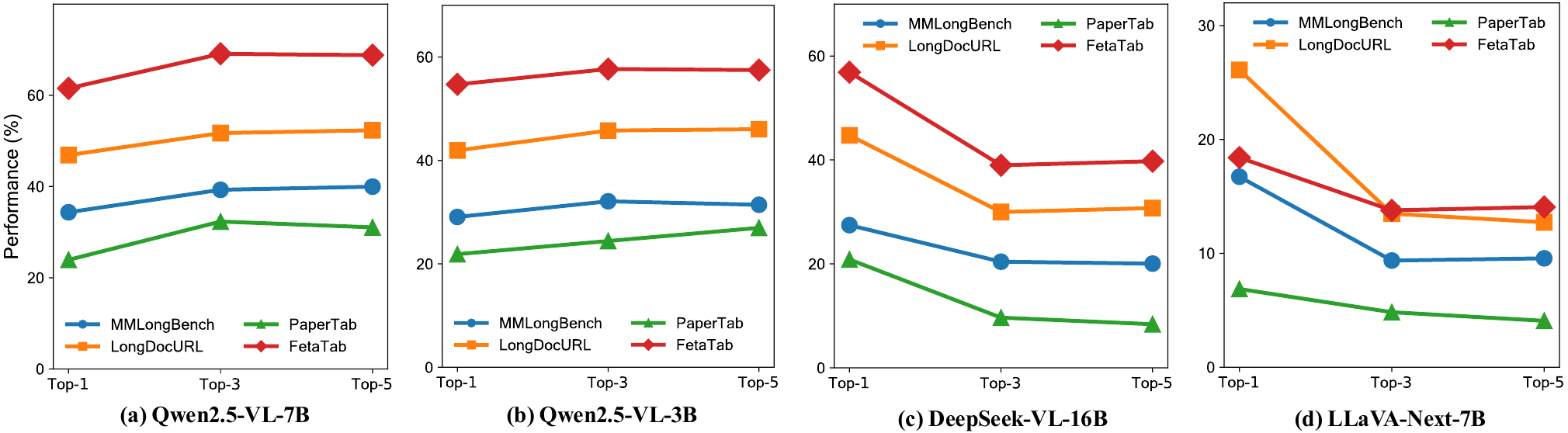}
    \caption{\textbf{Performance trends of MoLoRAG across different top-$K$ retrieval settings for LVLMs.} LVLMs with extensive context support (e.g., Qwen2.5-VL series) benefit from retrieving more pages, improving performance with higher $K$. In contrast, LVLMs with limited context capacity (e.g., LLaVA-Next-7B) perform best with $K=1$.}
    \label{fig:trend_with_k}
    \vspace*{-7pt}
\end{figure*}

\noindent \textbf{Choices of LLMs} For LLMs, we consider Mistral-7B-Instruct-v0.2 \cite{Jiang2023Mistral7B}, Qwen2.5-7B-Instruct \cite{qwen2025qwen25}, LLaMA3.1-8B-Instruct \cite{grattafiori2024llama}, GPT-4o, and DeepSeek-V3 \cite{deepseekai2025deepseekv3}. These models vary in series, scales, reasoning capabilities, and open-source availability, offering a diverse evaluation of LLM-based methods. 

\noindent \textbf{Choices of LVLMs} We classify LVLMs into three categories based on their input capacity: (1) Large Input Size models, such as Qwen2.5-VL-3B and Qwen2.5-VL-7B, which can process extensive context sizes, e.g., 30 images. (2) Medium Input Size models, such as DeepSeek-VL-16B \cite{lu2024deepseekvl}, which can handle a moderate number of inputs, e.g., 5 images. (3) Single Input models, such as LLaVA-Next-7B, which are limited to processing one image at a time. For each LVLM backbone, we assess its compatibility with various methods and sensitivity to context size, providing guidelines for effectively leveraging LVLMs in document understanding tasks.

\subsection{Overall Performance}
In this subsection, we present the overall performance of MoLoRAG alongside all baseline methods. To evaluate performance under varying retrieval availability, we consider top-$K$ values of $K=1, 3, 5$. Results for top-$3$ retrieval are shown in Table \ref{tab:main_top3}, while additional results for $K=1$ and $K=5$ are provided in Tables \ref{tab:main_top1} and \ref{tab:main_top5} in Appendix, respectively.  For LLM w. Text RAG, each retrieved element corresponds to a text chunk, whereas for LVLM-based methods, each retrieved element represents a  document page in image format. Based on the experimental results, we summarize the key findings below:

\noindent \textbf{1. LLMs struggle with document understanding compared to LVLM-based methods.} Even advanced LLMs like DeepSeek-V3, fall short in performance compared to LVLM-based methods. This highlights the inherent limitations of LLMs in handling multi-modal document understanding tasks, even when paired with sophisticated retrieval methods. LVLMs, on the other hand, can natively handle multi-modal inputs, making them better suited for document understanding. A fine-grained analysis across different evidence modalities (e.g., text, tables, figures) in Appendix \ref{appendix:fine_grained} reveals LLMs' weak performance with non-text modalities, while LVLMs excel across diverse modalities. 

\noindent \textbf{2. MoLoRAG consistently boosts LVLM performance.} Integrating LVLMs with MoLoRAG significantly improves their question answering capabilities. For example, DeepSeek-VL-16B, which performs poorly with concatenated document images (e.g., 8.40\% on MMLongBench due to content overload), achieves a substantial improvement when paired with MoLoRAG, reaching 20.43\%. Similarly, high-capacity LVLMs like the Qwen2.5-VL series benefit from MoLoRAG's ability to filter and prioritize relevant pages, further improving their already strong performance.

\noindent \textbf{3. Fine-tuned MoLoRAG+ delivers further performance gains.} The fine-tuned variant, MoLoRAG+, outperforms the training-free version, demonstrating the benefits of task-specific optimization. For example, with DeepSeek-VL-16B, MoLoRAG+ achieves 5.04\% improvement on MMLongBench compared to the training-free MoLoRAG. This enhancement stems from its superior ability to assess logical relevance, enabling more accurate retrieval (details in Section \ref{sec:retrieve_comp}). 

\input{tables/retrieve}

\begin{figure*}
    \centering
    \includegraphics[width=0.92\linewidth]{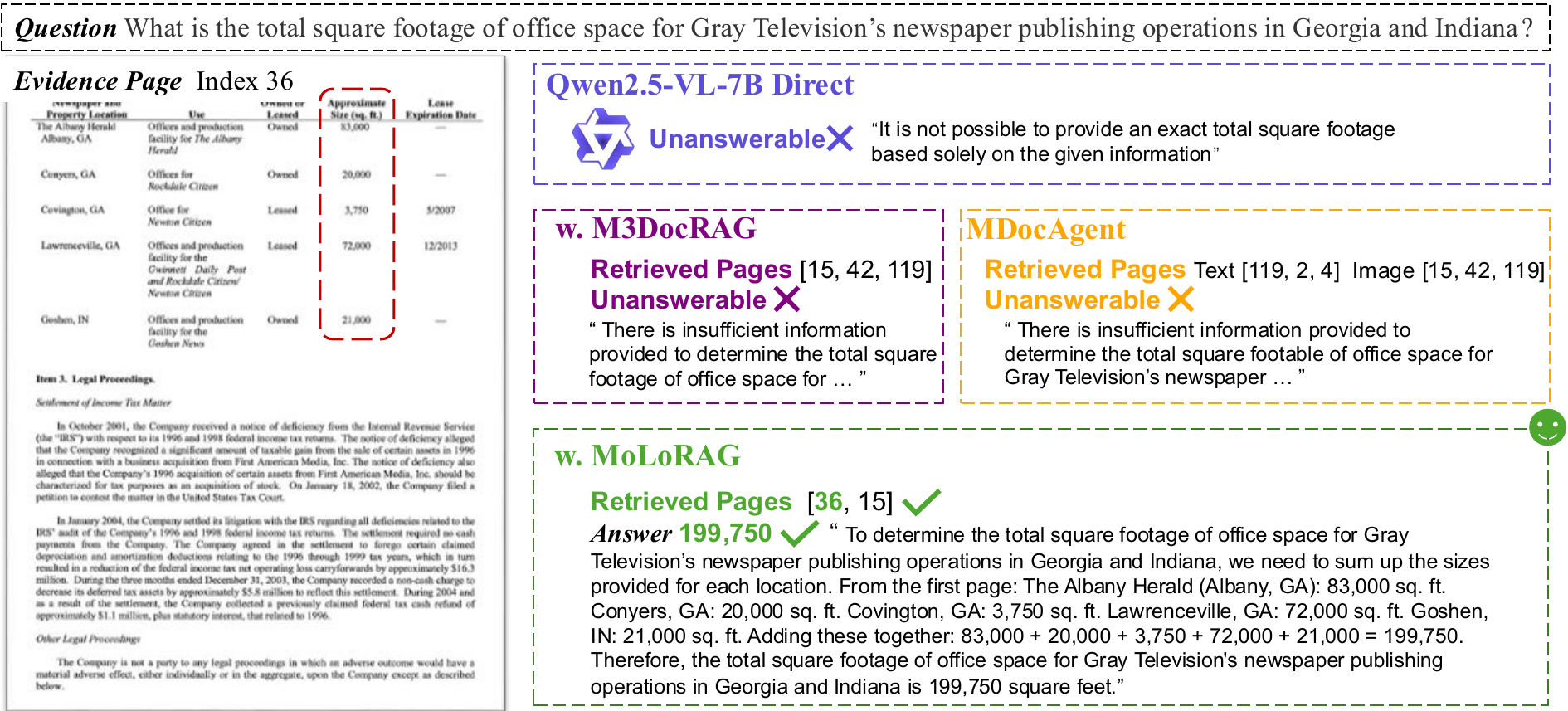}
    \caption{\textbf{Case study on LongDocURL.} MoLoRAG successfully \textbf{retrieves the correct evidence page} for the given question by leveraging logical relevance, enabling it to \textbf{provide the correct answer}. In contrast, both LVLM direct inference and other baseline methods fail to answer the question due to limited or irrelevant context.}
     \vspace*{-11pt}
    \label{fig:case_study}
\end{figure*}

\noindent \textbf{4. The relationship between top-$K$ retrieval and performance depends on LVLM capability.} Figure \ref{fig:trend_with_k} illustrates how performance of MoLoRAG varies with top-$K$ retrieval settings across four datasets. For LVLMs with extensive context support, such as the Qwen2.5-VL series, increasing $K$ (i.e., providing more pages) improves performance across all scenarios. However, for LVLMs with limited context capacity, such as DeepSeek-VL-16B and LLaVA-Next-7B, additional pages often exceed their processing capabilities, leading to degraded performance. For these models, $K=1$ is typically the optimal choice.

\subsection{Retrieval Performance Comparison}\label{sec:retrieve_comp}

In this subsection, we evaluate the retrieval accuracy of various methods to highlight the effectiveness of MoLoRAG in identifying relevant pages. Since only MMLongBench and LongDocURL provide ground-truth evidence pages for each query, our comparison is confined to these two datasets. We employ standard metrics, including Recall, Precision, NDCG, and MRR (details in Appendix \ref{appendix:metrics}), where higher values indicate better retrieval performance. We compare MoLoRAG and its fine-tuned variant, MoLoRAG+, against two baseline methods: M3DocRAG \cite{cho2024m3docrag} and MDocAgent \cite{han2025mdocagent}. MDocAgent performs separate text- and image-based retrieval, and results for both modalities are reported. The detailed results under top-$K$ settings are presented in Table \ref{tab:retrieval_comp}. MoLoRAG consistently outperforms baseline methods across metrics, with an average improvement of \textbf{9.94\%} on MMLongBench and \textbf{7.16\%} on LongDocURL. This advantage arises from MoLoRAG’s integration of both semantic and logical relevance, unlike the baselines, which focus solely on semantic relevance. The fine-tuned variant, MoLoRAG+, further improves performance by leveraging task-specific optimization.

\subsection{Case Study} 

Figure \ref{fig:case_study} presents a case study on LongDocURL. LVLM direct inference marks the question as ``unanswerable'' due to limited input context. Baselines such as M3DocRAG and MDocAgent rely solely on semantic relevance for retrieval, failing to locate the evidence page, which leads to incorrect answers. In contrast, MoLoRAG accurately retrieves the evidence page by considering logical relevance, enabling the LVLM to leverage this knowledge and correctly answer the question. Another case involving \textbf{cross-page understanding} is illustrated in Figure \ref{fig:case_study_2} in the Appendix. 

\noindent Due to space limits, \textbf{ablation study} and \textbf{efficiency analysis} are moved to Appendix \ref{appendix:ablation} and \ref{appendix:efficiency}.

%% file: tables/main_top3.tex
\begin{table*}[!t]
    \centering
     \caption{\textbf{Overall performance comparison (in $\%$) under the retrieved top-$3$ setting.} The ``Direct'' mode processes up to 30 document pages, while ``MoLoRAG+'' refers to the variant with a fine-tuned retrieval engine. Results for the top-$1$ and top-$5$ settings are in Tables \ref{tab:main_top1} and \ref{tab:main_top5}, respectively. The best performance is \colorbox{orange!20}{\textbf{highlighted}}.}
     \vspace*{-8pt}
    \resizebox{0.95\linewidth}{!}{

       \begin{tabular}{ccc|cccc|c}
      \toprule
      \rowcolor{COLOR_MEAN} \textbf{Type} & \textbf{Model} & \textbf{Method}  & \textbf{MMLongBench}  & \textbf{LongDocURL} & \textbf{PaperTab} & \textbf{FetaTab} & \textbf{Avg.} \\ \midrule 
      \multirow{5}{*}{\textit{\textbf{LLM-based}}} & Mistral-7B & Text RAG & 24.47 & 25.06 & 11.45 & 41.14 & 25.53 \\ 
     & Qwen2.5-7B & Text RAG & 25.52 & 27.93 & 12.72 & 40.06 & 26.56  \\ 
     & LLaMA3.1-8B & Text RAG & 22.56 & 29.80 & 13.49 & 45.96 & 27.95  \\ 
     & GPT-4o & Text RAG & 27.23 & 32.74 & 14.25 & 50.20 & 31.11  \\ 
     & DeepSeek-V3 & Text RAG & \textbf{29.82} & \textbf{34.73} & \textbf{17.05} & \textbf{52.36} & \textbf{33.49} \\  \midrule 

     \multirow{16}{*}{\textit{\textbf{LVLM-based}}} & \multirow{4}{*}{LLaVA-Next-7B} & Direct & 7.15 & 10.78 & 3.05 & 11.61 & 8.15 \\ 
     & & M3DocRAG & \textbf{10.10} & \textbf{13.85} & 5.34 & \textbf{13.98} & \textbf{10.82}  \\ 
     & & MoLoRAG & 9.37 & 13.49 & 4.83 & 13.78 & 10.37 \\ 
     & & MoLoRAG+ & 9.47 & 13.58 & \textbf{5.60} & 13.48 & 10.53 \\ \cmidrule{2-8}

      & \multirow{4}{*}{DeepSeek-VL-16B} & Direct & 8.40 & 14.72 & 6.11 & 16.14 & 11.34 \\ 
     &  & M3DocRAG & 18.12 & 29.60 & 7.89 & 27.07 & 20.67 \\ 
     &  & MoLoRAG & 20.43 & 29.98 & 9.67 & 38.98 & 24.77 \\ 
     &  & MoLoRAG+ & \textbf{25.47} & \textbf{37.21} & \textbf{10.94} & \textbf{41.54} & \textbf{28.79} \\ \cmidrule{2-8}
     
       & \multirow{4}{*}{Qwen2.5-VL-3B} & Direct & 26.65 & 24.89 & 25.19 & 51.57 & 32.08 \\ 
     &  & M3DocRAG &  29.11 & 44.40 & 24.68 & 53.25 & 37.86 \\
     &  & MoLoRAG & 32.11 & \textbf{45.79} & 24.43 & 57.68 & 40.00 \\ 
     &  & MoLoRAG+ & \textbf{32.47} & 45.27 & \textbf{27.23} & \textbf{58.76} & \textbf{40.93}  \\ \cmidrule{2-8}
        
     &  \multirow{4}{*}{Qwen2.5-VL-7B} & Direct & 32.77 & 26.38 & 29.77 & 64.07 & 38.25 \\ 
     & & M3DocRAG & 36.18 & 49.03 & 28.50 & 63.78 & 44.37 \\ 
     & & MoLoRAG & 39.28 & 51.71 & \cellcolor{orange!20} \textbf{32.32} & 69.09 & 48.10 \\ 
     & & MoLoRAG+ & \cellcolor{orange!20} \textbf{41.01} & \cellcolor{orange!20}\textbf{51.85} & 31.04 & \cellcolor{orange!20} \textbf{69.19} & \cellcolor{orange!20} \textbf{48.27} \\ \midrule 

     \textit{\textbf{Multi-agent}} & \multicolumn{2}{c|}{MDocAgent (LLaMA3.1-8B+Qwen2.5-VL-7B)} & 38.53 & 46.91 & 30.03 & 66.34 & 45.45 \\ 
       
      \bottomrule
    \end{tabular}

    }
    \label{tab:main_top3}
\end{table*}

%% file: tables/retrieve.tex
\begin{table*}[!t]
    \centering
    \caption{\textbf{Retrieval performance comparison (in $\%$) under the top-$K$ setting.}}
    \label{tab:retrieval_comp}
    \vspace*{-8pt}
    \resizebox{0.82\linewidth}{!}{

     \begin{tabular}{cc|cccc|cccc} 
      \toprule
      \rowcolor{COLOR_MEAN}  & & \multicolumn{4}{c|}{\textbf{MMLongBench}} & \multicolumn{4}{c}{\textbf{LongDocURL}} \\
        \rowcolor{COLOR_MEAN} \multirow{-2}{*}{\textbf{Top-$K$}} & \multirow{-2}{*}{\textbf{Method}} & \textbf{Recall} & \textbf{Precision} & \textbf{NDCG} & \textbf{MRR}  & \textbf{Recall} & \textbf{Precision} & \textbf{NDCG} & \textbf{MRR}   \\ \midrule 
       \multirow{5}{*}{$1$} & M3DocRAG & 43.31 & 56.67 & 56.67 & 56.67 & 46.84 & 64.66 & 64.66 & 64.66 \\ 
        & MDocAgent (Text) & 29.30 & 38.99 & 38.99 & 38.99 & 42.03 & 58.37 & 58.37 & 58.37 \\ 
        & MDocAgent (Image) & 43.79 & 57.49 & 57.49 & 57.49 & 46.80 & 64.57 & 64.57 & 64.57 \\ 
       \rowcolor{blue!10} \cellcolor{white} & MoLoRAG & 45.46 & 59.95 & 59.95 & 59.95 & 48.98 & 67.71 & 67.71 & 67.71 \\ 
       \rowcolor{blue!10} \cellcolor{white} & MoLoRAG+ & \textbf{51.32} & \textbf{66.86} & \textbf{66.86} & \textbf{66.86}  & \textbf{50.82} & \textbf{70.08} & \textbf{70.08} & \textbf{70.08} \\ \midrule 

       \multirow{5}{*}{$3$} & M3DocRAG & 64.17 & 31.62 & 54.13 & 65.36 & 67.00 & 33.78 & 58.23 & 72.51  \\ 
       & MDocAgent (Text) & 43.21 & 20.77 & 37.13 & 45.26 & 58.53 & 29.33 & 54.12 & 65.28 \\ 
       & MDocAgent (Image) & 64.74 & 31.97 & 54.75 & 66.12 & 66.67 & 33.62 & 58.26 & 72.47 \\ 
       \rowcolor{blue!10} \cellcolor{white} & MoLoRAG & 67.22 & 40.81 & 57.34 & 68.56 & \textbf{70.04} & 36.41 & 61.56 & 75.78 \\ 
       \rowcolor{blue!10} \cellcolor{white} & MoLoRAG+ & \textbf{68.87} & \textbf{48.67} & \textbf{64.49} & \textbf{73.50} & 68.92 & \textbf{47.53} & \textbf{64.90} & \textbf{77.14} \\ \midrule
       
        \multirow{5}{*}{$5$} & M3DocRAG & 72.00 & 22.58 & 54.06 & 66.92 & 74.32 & 23.34 & 58.05 & 73.83 \\ 
        & MDocAgent (Text) & 50.60 & 15.48 & 37.19 & 46.98 & 65.41 & 20.41 & 53.97 & 66.55 \\ 
        & MDocAgent (Image) & 71.45 & 22.37 & 54.58 & 67.53 & 74.60 & 23.50 & 58.06 & 73.90  \\ 
       \rowcolor{blue!10} \cellcolor{white} & MoLoRAG & \textbf{74.13} & 35.83 & 57.29 & 69.63 & \textbf{77.14} & 26.13 & 61.30 & 76.88 \\ 
       \rowcolor{blue!10} \cellcolor{white} & MoLoRAG+  & 72.37 & \textbf{45.34} & \textbf{64.36} & \textbf{73.97} & 73.69 & \textbf{42.47} & \textbf{64.74} & \textbf{77.89} \\  \bottomrule
        
    \end{tabular}
    
    }
    
\end{table*}

%% file: appendix/1-pseudo-code.tex
\section{Algorithm Pseudo-Code}\label{appendix:algorithm}

\begin{algorithm}[!t]
\caption{\textbf{Graph Traversal for Retrieval}}
\label{algorithm:beamsearch}
\begin{algorithmic}[1]
\Require Question $q$, $\text{DocEncoder}(\cdot)$, Page graph $G$, Exploration size $w$, Hop limit $n_{\text{hop}}$, Document $\mathcal{D}$
\Ensure Re-ranked pages $\mathcal{D}^r$

\State $s^{\text{sem}}_i \gets \langle  \text{DocEncoder}(q), \text{DocEncoder}(p_i) \rangle $ for each $p_i \in \mathcal{D}$ \Comment{\textcolor{red}{\small{Semantic relevance scoring}}}
\State $\mathcal{B} \gets \text{TopK}( \{ s^{\text{sem}}_i \}, w)$ \Comment{\textcolor{red}{\small{Exploration set initialization}}}
\State $\mathcal{D}^r \gets \emptyset$ 
\State $\mathcal{S} \gets \mathcal{B}$ \Comment{\textcolor{red}{\small{Visited marking}}}
\For{$p_i \in \mathcal{B}$}
    \State $s^{\text{logi}}_i \gets \text{VLM}(q, p_i)$ \Comment{\textcolor{red}{\small{Logical relevance scoring}}} 
    \State $s_i \gets \text{Combine}(s_i^{\text{sem}}, s_i^{\text{logi}})$ \Comment{\textcolor{red}{\small{Score update}}}
    \State $\mathcal{D}^r \gets \mathcal{D}^r \cup \{(p_i, s_i)\}$
\EndFor

\For{$\text{Hop} = 1$ to $n_{\text{hop}}$}
    \State $\mathcal{C} \gets \emptyset$ \Comment{\textcolor{red}{\small{Candidates initialization}}}
    \For{$p_i \in \mathcal{B}$}
        \For{$p_j \in \text{Neighbor}(G, p_i), p_j \notin \mathcal{S}$} 
            \State $s_j^{\text{logi}} \gets \text{VLM}(q, p_j)$
            \State $s_j \gets \text{Combine}(s_j^{\text{sem}}, s_j^{\text{logi}})$
            \State $\mathcal{C} \gets \mathcal{C} \cup \{ s_j\}$
            \State $\mathcal{D}^r \gets \mathcal{D}^r \cup \{(p_j, s_j)\}$ 
            \State $\mathcal{S} \gets \mathcal{S} \cup \{ p_j \}$
        \EndFor
    \EndFor
    \State $\mathcal{B} \gets \text{TopK}(\mathcal{C}, w)$ \Comment{\textcolor{red}{\small{Exploration set update}}}
\EndFor

\State Sort $\mathcal{D}^r$ by descending $s$ \Comment{\textcolor{red}{\small{Pages re-ranking}}}
\State \Return $\mathcal{D}^r$
\end{algorithmic}
\end{algorithm}

The graph traversal algorithm for retrieval is presented in Algorithm \ref{algorithm:beamsearch}. This algorithm operates by efficiently identifying relevant pages through a combination of semantic and logical relevance scores. By leveraging an exploration size $w$ and a hop limit $n_{\text{hop}}$, the traversal is restricted to exploring only the most promising paths in the page graph, ensuring scalability and avoiding the need to process all pages. The output, $\mathcal{D}^r$, is a re-ranked set of document pages that are both semantically and logically relevant, with its size typically smaller than the total number of document pages $N$. From this re-ranked set, the top-$K$ pages are selected and passed to the next stage for question answering.

%% file: appendix/2-prompt.tex
\section{Prompt}\label{appendix:prompt}
In this section, we present all the prompts used within the MoLoRAG framework, including querying the VLM to assign logical relevance scores, using LLMs or LVLMs for question answering, and prompting GPT-4o to curate the dataset for training MoLoRAG+.

\noindent \textbf{Assessing Logical Relevance} The prompt for querying the VLM to assign a logical relevance score between the observed image and the question is provided below:
\begin{tcolorbox}[     
     colback=white, 
    colframe=titlepink,              
    coltitle=white,                   
    colbacktitle=titlepink,                  
    title=\textbf{Prompt for Assessing Logical Relevance} 
]
   
    \textbf{\# GOAL \# }
    
    You are an Retrieval Expert, and your task is to evaluate how \textbf{relevant} the input document page is to the given query. Rate the relevance on a scale of 1 to 5, where:

     \textbf{5} Highly relevant - contains complete information needed to answer the query
     
     \textbf{4}  Very relevant - contains most of the information needed
     
     \textbf{3}  Moderately relevant - contains some useful information
     
     \textbf{2}  Slightly relevant - has minor connection to the query
     
     \textbf{1} Irrelevant - contains no information related to the query

     \vspace*{5pt}
     
    \textbf{\# INSTRUCTION \#}
    
    Please first read the given query, think about what knowledge is required to answer that query, and then carefully go through the document snapshot for judgment. 

    \vspace*{5pt}
    
    \textbf{ \# QUERY \# }
    
     $\{\{$\texttt{Question}$\}\} $
     
     Please generate just a single number (1-5) representing your relevance judgment. Your answer should be a single number without any extra contents. 
\end{tcolorbox}

\noindent \textbf{LLM Question Answering} For LLM-based question answering, all retrieved text chunks are concatenated into the \texttt{context}, and the LLM is queried to answer the given question based on the provided context as:`` $\{ \{ \texttt{Context} \} \}$ Answer the question based on the above context:  
$\{ \{ \texttt{Question} \} \}$''. 

\noindent \textbf{LVLM Question Answering} For LVLM-based question answering, the input includes document images and the question: ``$\{ \{ \texttt{Image} \} \}$ Based on the document, please answer the question:   $\{ \{ \texttt{Question} \} \}$''.

\noindent \textbf{Curating Training Data} The prompt for  guiding GPT-4o to generate training data is shown below: 
\begin{tcolorbox}[     
     colback=white, 
    colframe=titlepink,              
    coltitle=white,                   
    colbacktitle=titlepink,                  
    title=\textbf{Prompt for Curating Training Data} 
]
   
    \textbf{\# GOAL \# }

    Given the input image, your task is to generate a question related to it. The relevance score is $\{ \{\texttt{relevance}\_\texttt{score} \} \}$, where a higher score indicates a closer connection between the question and the image. For example, a relevance score of 5 means the answer is \textbf{DIRECTLY} contained in the image, while a score below 3 indicates that the answer \textbf{CANNOT} be derived from it, with lower scores signifying less relevance.

    \vspace*{5pt}

    \textbf{\# REQUIREMENT \#} 
    
    The question must be based on the content of the input image, except when the relevance score is $\leq$ 2. For relevance scores of 4 or higher, create clear and straightforward questions with answers that are explicitly present in the image. For relevance scores of 3, generate questions that may require some inference but are still somewhat related to the content. For relevance scores of 2 or lower, formulate questions that are unanswerable based on the snapshot. 

   You may consider various elements, including text, layout, and figures. For this generation, please concentrate on $\{ \{\texttt{focus}\} \} $ if applicable and remember that the relevance score is $\{ \{\texttt{relevance}\_\texttt{score} \} \}$.

     \vspace*{5pt}

    Your output should be formatted as follows:
    \{ ``query'': ``Your generated question'', ``relevance\_score'': ``relevance\ score'', ``answer'': ``Corresponding answer or inference'' \} 
   
\end{tcolorbox}

%% file: appendix/3-train_star.tex
\section{Supplementary Materials for MoLoRAG+}\label{appendix:fine_tune_mmgraph}
This section provides detailed information on the training data, learning configurations, and alternative data engine for MoLoRAG+.

\noindent \textbf{Training Data} In the first stage, \textbf{Question Generation}, we prompt GPT-4o to generate approximately 5,500 samples using the illustrated prompt. Document snapshots are randomly selected from MMLongBench \cite{ma2024mmlongbench} and LongDocURL \cite{deng2024longdocurl}, as these datasets contain multi-modal, information-rich documents. The relevance score $s$ is sampled from $\{ 1, 2,3, 4,5\}$ with equal distribution to ensure a balanced representation across different levels of logical relevance, preventing over-fitting to specific scores. For each sampled document snapshot $p_i$, the generated question is expressed as: $q' = \text{GPT-4o}(\text{prompt}, s, p_i)$, where $p_i$ denotes the randomly selected document snapshot, and $s$ is the sampled relevance score. To ensure data quality, each generated question $q'$ and its corresponding document snapshot $p_i$ are fed back to GPT-4o to assign a predicted logical relevance score as: $s' = \text{GPT-4o}(\text{prompt}, q', p_i)$. Samples are retained only if the predicted score closely matches the original score, i.e., $|s - s'| \leq 1$. After this automated filtering, the remaining samples undergo manual verification. This process results in a final high-quality training set containing 3,519 samples, formatted as triplets: $\langle \text{Question} \; q', \text{Image} \; p_i, \text{Relevance Score} \; s' \rangle$. In each sample, $s'$, the predicted relevance score, is the expected output for model training. 

\noindent \textbf{Learning Configuration} We utilize the LLaMA-Factory package\footnote{\href{https://github.com/hiyouga/LLaMA-Factory/}{https://github.com/hiyouga/LLaMA-Factory/}} to fine-tune the backbone VLM, Qwen2.5-VL-3B \cite{Qwen2.5-VL}, using the LoRA technique for parameter-efficient training. The fine-tuning process is configured with the following hyperparameters: a LoRA Rank of 8, a learning rate of $1\times 10^{-4}$, a warmup ratio of 0.1, and gradient accumulation steps set to 8. The model is trained for a total of 2 epochs, ensuring efficient and effective parameter adaptation.

\noindent \textbf{Alternative Data Engine} We primarily use GPT-4o as the data engine for curating training data. However, our data generation pipeline is flexible and \textbf{can accommodate other LVLMs}. To demonstrate this flexibility, we replace the original engine with the open-source model Qwen2.5-VL-32B \cite{Qwen2.5-VL}, while keeping all prompts and processes consistent. This variant is denoted as MoLoRAG$^{+}_{\text{Qwen}}$. We compare the retrieval performance of MoLoRAG, MoLoRAG+, and MoLoRAG$^{+}_{\text{Qwen}}$ on MMLongBench (Table \ref{tab:molorag_qwen}), where the backbones are pre-trained Qwen2.5-VL-3B, Qwen2.5-VL-3B fine-tuned with GPT-4o data, and fine-tuned with Qwen2.5-VL-32B data, respectively. The results indicate that the Qwen2.5-VL-32B distilled model performs \textbf{comparably} to its GPT-4o counterparts. This is attributed to (1) the simplicity of logical relevance scoring, enabling effective high-quality data generation by Qwen2.5-VL-32B, and (2) the shared family of the data engine and distilled model, which facilitates capability transfer. Additionally, this variant reduces data construction costs due to its open-source nature, highlighting the effectiveness of our data construction pipeline.

\begin{table}[!t]
    \centering
    \caption{\textbf{Retrieval performance comparison (in $\%$)} between MoLoRAG, MoLoRAG$^+_{\text{Qwen}}$, and MoLoRAG+ on MMLongBench.}
    \resizebox{\linewidth}{!}{
    \begin{tabular}{cc|cccc}
      \toprule 
     \rowcolor{COLOR_MEAN} \textbf{Top-$K$} & \textbf{Model} & \textbf{Recall} & \textbf{Precision} & \textbf{NDCG} & \textbf{MRR} \\ 
      \midrule
      \multirow{3}{*}{$1$}   &  MoLoRAG & 45.46 & 59.95 & 59.95 & 59.95 \\ 
      & MoLoRAG$^+_{\text{Qwen}}$ & \textbf{51.62} & \textbf{67.56} & \textbf{67.56} & \textbf{67.56} \\ 
         & MoLoRAG+ & 51.32 & 66.86 & 66.86 & 66.86 \\ \midrule
         
      \multirow{3}{*}{$3$} & MoLoRAG & 67.22 & 40.81 & 57.34 & 68.56 \\ 
      & MoLoRAG$^+_{\text{Qwen}}$ & \textbf{71.79} & 37.94 & 64.24 & \textbf{74.57}  \\ 
      & MoLoRAG+ &  68.87 & \textbf{48.67} & \textbf{64.49} & 73.50 \\  \midrule 
      \multirow{3}{*}{$5$} & MoLoRAG & 74.13 & 35.83 & 57.29 & 69.63 \\ 
      & MoLoRAG$^+_{\text{Qwen}}$ & \textbf{78.72} & 29.06 & 64.01 & \textbf{75.69} \\ 
      & MoLoRAG+ &  72.37 & \textbf{45.34} & \textbf{64.36} & 73.97 \\ 
    \bottomrule
    \end{tabular}
    }
    \label{tab:molorag_qwen}
\end{table}

%% file: appendix/4-supple_experiment.tex
\section{Supplementary Materials for Experiments} 

\subsection{Details of Metrics}\label{appendix:metrics}

\textbf{Evaluation for Question Answering} We utilize \textbf{Accuracy} and \textbf{Exact Match} as evaluation metrics for MMLongBench \cite{ma2024mmlongbench} and LongDocURL \cite{deng2024longdocurl}. Accuracy is rule-based to accommodate various answer types, with detailed explanations provided in Appendix B.3 of \citet{ma2024mmlongbench}. Exact Match measures the percentage of predictions where the generated answer \textbf{exactly} matches the ground-truth answer. For the remaining datasets, PaperTab and FetaTab \cite{hui2024uda}, we employ GPT-4o as the evaluator to assign a \textbf{Binary Correctness} score $\in \{ 0, 1 \}$, where $1$ indicates that the generated answer aligns with the ground-truth answer. We report the averaged values across all test samples. 

\noindent \textbf{Evaluation for Retrieval} We employ standard retrieval metrics, including Recall@$K$, Precision@$K$, NDCG@$K$, and MRR@$K$, where $K$ represents the number of retrieved elements. For a specific data sample with ground-truth evidence pages denoted as $\mathcal{P}^{\text{gt}} = \{ p_1^{\text{gt}}, \ldots, p_n^{\text{gt}} \}$ and the top-$K$ retrieved pages $\mathcal{P}^r = \{ p_1^{r}, \ldots, p_K^r \}$, these metrics are computed as follows:

\begin{itemize}
    \item \textbf{Recall} measures the proportion of ground-truth pages that are successfully retrieved within the top-$K$ results:

    $$\text{Recall}@K = \frac{\sum_{i=1}^K \mathbb{I}(p_i^r \in \mathcal{P}^{\text{gt}})}{n},$$

    where $\mathbb{I}(\cdot)$ is the indicator function that returns $1$ if the condition is true and $0$ otherwise. 

    \item \textbf{Precision} assesses the accuracy of the retrieved pages by calculating the proportion of retrieved pages that are relevant:
    
    $$\text{Precision}@K = \frac{\sum_{i=1}^K \mathbb{I}(p_i^r \in \mathcal{P}^{\text{gt}}) }{K}.$$ 

    \item \textbf{NDCG} (Normalized Discounted Cumulative Gain) evaluates the ranking quality of the retrieved pages by considering the positions of relevant pages within the top-$K$ results. It is computed in three steps:

    $$ \text{DCG}@K = \sum_{i=1}^{\min(n, K)} \frac{\mathbb{I}(p_i^r \in \mathcal{P}^{\text{gt}})}{\log_2 (i + 1)}, $$

    $$\text{IDCG}@K = \sum_{i=1}^{\min(n, K)} \frac{1}{\log_2 (i + 1)}, $$

    $$ \text{NDCG}@K = \frac{\text{DCG}@K}{\text{IDCG}@K}$$

    \item \textbf{MRR} (Mean Reciprocal Rank) measures the reciprocal rank of the first relevant page within the top-$K$ retrieved pages. It is defined as:
    \[
    \text{MRR}@K = 
    \begin{cases}
        \frac{1}{i} & \text{if } p_i^{r} \text{ is the \textbf{first} relevant page} \\
        0 & \text{otherwise}
    \end{cases}
    \]

    where $i$ denotes the position of the \textbf{first relevant page} that satisfies $p_i^r \in \mathcal{P}^{\text{gt}}$. If no relevant page is retrieved within the top-$K$, the MRR score for that sample is $0$.
\end{itemize}

For each metric, we average the scores over all test samples and report the values in Tables \ref{tab:retrieval_comp}, \ref{tab:ablation_logic} and \ref{tab:ablation_fulldoc}, respectively.

\subsection{Implementation Details}\label{appendix:implementation}

This subsection outlines the detailed configurations for each method to ensure clarity and reproducibility. All experiments were conducted on 3 NVIDIA A6000 48G GPUs.

\begin{itemize}
    \item \textbf{LLM w. Text RAG}  We use the \texttt{PyPDFLoader} from the LangChain package\footnote{\texttt{\href{https://python.langchain.com/}{https://python.langchain.com/}}} to extract text from the raw document. Each document is divided into chunks of 1,000 tokens. The QwenRAG API\footnote{\texttt{\href{https://dashscope.console.aliyun.com/overview}{https://dashscope.console.aliyun.com/overview}}} is employed as the text RAG engine. Specifically, we use the \texttt{text-embedding-v1} model to encode text chunks and perform retrieval. For each top-$K$ setting, the top-ranked $K$ text chunks are combined as context, which is then passed to the LLM for answering. 

    \item \textbf{LVLM Direct Inference} To ensure scalability, we truncate documents to retain only the first 30 pages. For Qwen2.5-VL series, as these models support extensive image contexts, all images are fed directly for processing. For DeepSeek-VL-16B, although it supports multi-image inputs, it requires significant memory for loading. Therefore, we concatenate document images into 5 larger images, ensuring compatibility with GPU memory. For LLaVA-Next-7B, as it accepts only a single image, all available pages are combined into one single image for processing. 

    \item \textbf{M3DocRAG} This baseline is implemented according to its original paper and official repository. The document encoder used is \texttt{colpali} for encoding and retrieval. While the original paper uses Qwen2-VL-7B \cite{wang2024qwen2vl} as the backbone LVLM, we extend the evaluation by integrating the method with various LVLMs to assess compatibility. 

    \item \textbf{MDocAgent} This baseline is implemented following its official repository: \texttt{colpaligemma-3b-mix-448-base} for image retrieval and \texttt{colbertv2.0} for text retrieval. For all five agents in this framework, we consistently use the original LLaMA3.1-8B \cite{grattafiori2024llama} as the LLM for the text agent, while employing a consistent LVLM, i.e., Qwen2.5-VL-7B \cite{Qwen2.5-VL}, for remaining agents. 

    \item \textbf{MoLoRAG} For document encoding, \texttt{colpali} is used as the document encoder. Pages are connected in the graph if their similarity score exceeds a threshold of 0.4. During graph traversal, the number of hops $n_{\text{hop}}$ is set to 4, and the exploration set size $w$ is set to 3. Semantic relevance and LVLM-generated logical relevance are combined using an average score. All visited pages are re-ranked based on this combined score for final retrieval. Qwen2.5-VL-3B is employed as the retrieval engine due to its strong performance and lightweight architecture, ensuring both effectiveness and efficiency. The retrieved top-$K$ images are fed into LVLMs based on their input format capabilities: for LLaVA-Next-7B, the images are concatenated into a single composite image, while for all other LVLMs, the images are processed separately. 

\end{itemize}

\subsection{Discussion of Advanced OCR Methods}\label{appendix:advanced_ocr} 

We expand our discussion on the LLM w. Text RAG baseline by using more advanced OCR tools, including \texttt{MinerU} \cite{MinuerU} and \texttt{GOT-OCR-2.0} \cite{GOT-OCR}, as alternatives to \texttt{PyPDFLoader}, while ensuring consistency in the retrieval engine and LLM calling process. The results across different top-$K$ settings on MMLongBench and PaperTab are presented in Table \ref{tab:advanced_ocr}. Our findings indicate that advanced tools generally enhance QA performance due to improved OCR capabilities, with the benefits becoming more pronounced as $K$ increases. For instance, replacing \texttt{PyPDFLoader} with \texttt{MinerU} yields performance gains of up to 4\% across various LLMs. However, \textbf{LLMs w. Text RAG baselines still show a performance gap compared to strong LVLMs}, primarily due to the inevitable loss of multi-modal information.

\begin{table*}[!t]
    \centering
     \caption{\textbf{Comparison of various OCR methods on the performance of LLM w. Text RAG.}}
   \resizebox{0.95\linewidth}{!}{
    \begin{tabular}{cc|cc|cc|cc}
      \toprule 
     \rowcolor{COLOR_MEAN} & & \multicolumn{2}{c|}{\textbf{Top-$1$}} & \multicolumn{2}{c|}{\textbf{Top-$3$}} & \multicolumn{2}{c}{\textbf{Top-$5$}} \\ 
    \rowcolor{COLOR_MEAN} \multirow{-2}{*}{\textbf{Model}} & \multirow{-2}{*}{\textbf{OCR}} & \textbf{MMLongBench} & \textbf{PaperTab} & \textbf{MMLongBench} & \textbf{PaperTab} & \textbf{MMLongBench} & \textbf{PaperTab}  \\ \midrule 
       \multirow{3}{*}{\textbf{Qwen2.5-7B}} & \texttt{PyPDFLoader} & 22.11 & 5.34 & 25.52 & 12.72 & 26.09 & 16.79 \\  
       & \texttt{GOT-OCR-2.0} & 22.66 & 6.11  & 25.84 & 13.74 & 26.70 & 16.79  \\ 
       & \texttt{MinerU} & 21.99 & 7.63 & 24.54 & 16.28 & 27.53 & 19.08 \\  \midrule

      \multirow{3}{*}{\textbf{GPT-4o}} & \texttt{PyPDFLoader} & 24.07 & 8.65 & 27.23 & 14.25 & 28.74 & 20.36 \\ 
      & \texttt{GOT-OCR-2.0} & 23.86 & 8.91 & 27.78 & 13.74 & 29.47 & 18.07 \\ 
      & \texttt{MinerU} & 25.89 & 9.92 & 28.74 & 18.32 & 30.98 & 22.90  \\ \midrule 

      \multirow{3}{*}{\textbf{DeepSeek-V3}} & \texttt{PyPDFLoader} & 25.94 & 10.18  & 29.82 & 17.05  & 31.23 & 23.92 \\ 
      & \texttt{GOT-OCR-2.0} & 25.35 & 9.92 & 28.39 & 18.83 & 30.75 & 22.90 \\ 
      & \texttt{MinerU} & 26.21 & 11.96 & 30.11 & 21.12  & 32.36 & 26.97\\ \midrule

      \multicolumn{2}{c|}{\textbf{Qwen2.5-VL-7B w. MoLoRAG}} & 34.35 & 23.92 & 39.28 & 32.32 & 39.97 & 31.04 \\
    \bottomrule
    \end{tabular}
   }
    \label{tab:advanced_ocr}
\end{table*}

\subsection{Efficiency Analysis}\label{appendix:efficiency}
In this subsection, we provide efficiency analysis of our MoLoRAG with baseline methods from three aspects: (1) Retrieval scalability, (2) Inference efficiency, and (3) Total time costs. 
\begin{figure}[!t]
    \centering
    \includegraphics[width=\linewidth]{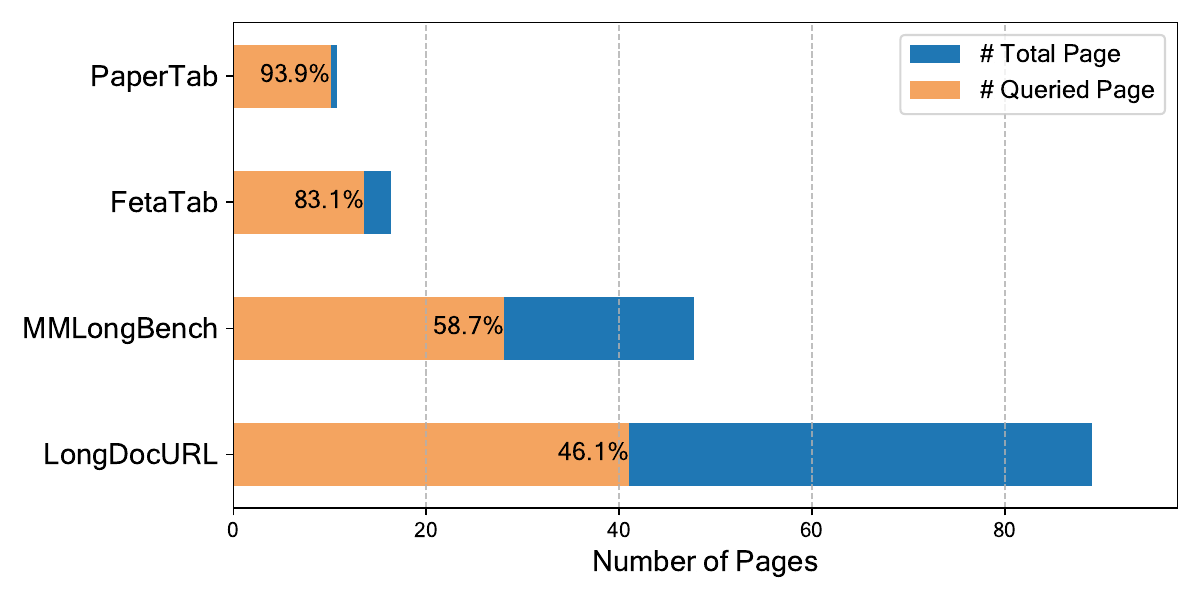}
    \caption{\textbf{Illustration of retrieval scalability.} We present the average number of pages queried by MoLoRAG and the total number of pages across all test samples for each dataset. }
    \label{fig:retrieval_scalability}
\end{figure}

\noindent \textbf{Retrieval Scalability} The retrieval stage of MoLoRAG requires the VLM to evaluate the logical relevance score of each visited page. To efficiently manage the traversal scope, we introduce hyper-parameters such as hop limit and exploration set size, which help narrow the querying space, ensuring scalability and accelerating the retrieval process. To illustrate the scalability of MoLoRAG, Figure \ref{fig:retrieval_scalability} shows the average number of queried pages alongside the total number of pages for all testing samples across different datasets. The figure also indicates the percentage of queried pages. As the document size increases, the average percentage of queried pages decreases significantly. For instance, fewer than 50\% of the pages are queried for LongDocURL, demonstrating MoLoRAG's ability to effectively control graph traversal and focus on relevant pages. This reduction highlights the scalability of the method, as it maintains high retrieval accuracy while minimizing computational overhead even for large documents.

\begin{figure}[!t]
    \centering
    \includegraphics[width=0.82\linewidth]{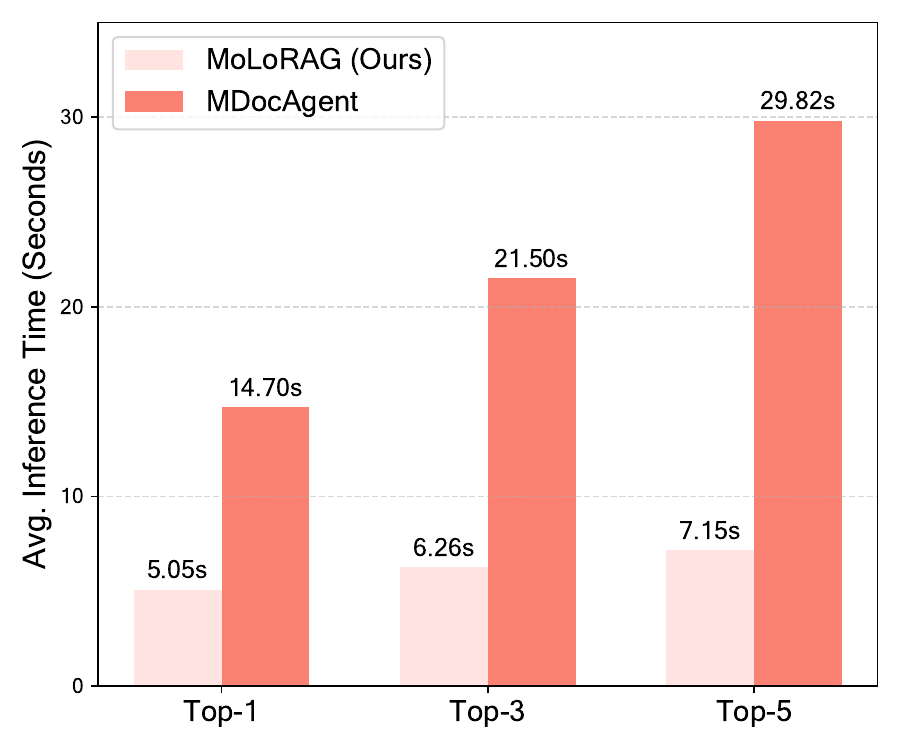}
    \caption{\textbf{Inference time comparison between MoLoRAG and MDocAgent \cite{han2025mdocagent} on the MMLongBench.}}
    \label{fig:inference_time}
\end{figure}

\noindent \textbf{Inference Efficiency} After retrieval, MoLoRAG requires only \textbf{a single query} to the LVLM for question answering by providing the relevant pages, ensuring efficiency comparable to LVLM Direct Inference. In contrast, the best-performing baseline, MDocAgent \cite{han2025mdocagent}, requires \textbf{five separate queries} to both LLMs and LVLMs, as it functions as a unified multi-agent system. Figure \ref{fig:inference_time} illustrates the average inference times of MoLoRAG (Qwen2.5-VL-7B as the backbone) and MDocAgent, clearly highlighting MoLoRAG’s significant advantage in inference efficiency, making it a more practical and scalable solution.

\begin{table*}[!t]
 \caption{\textbf{Total time costs comparison (in seconds) across different methods on LongDocURL.}}
    \centering
    \resizebox{\linewidth}{!}{
    \begin{tabular}{ccc|ccc|ccc}
      \toprule
        \rowcolor{COLOR_MEAN} & & & \multicolumn{3}{c|}{\textbf{Top-$1$}} &  \multicolumn{3}{c}{\textbf{Top-$5$}}  \\
        \rowcolor{COLOR_MEAN} \multirow{-2}{*}{\textbf{Type}} & \multirow{-2}{*}{\textbf{Model}}  & \multirow{-2}{*}{\textbf{Method}} & \textbf{Retrieval Time} & \textbf{Inference Time} & \textbf{Total} &  \textbf{Retrieval Time} & \textbf{Inference Time} & \textbf{Total} \\ \midrule 
        \multirow{5}{*}{\textbf{\textit{LLM-based}}} & Mistral-7B & Text RAG & 48.5s & 2.4s & 50.9s & 48.5s & 4.1s & 52.6s \\ 
        & Qwen2.5-7B & Text RAG & 48.5s & 5.1s & 53.6s & 48.5s & 6.9s & 55.4s \\ 
        & LLaMA3.1-8B & Text RAG & 48.5s & 11.8s & 60.3s & 48.5s & 18.3s & 66.8s \\ 
        & GPT-4o & Text RAG & 48.5s & 1.4s & 49.9s & 48.5s & 1.5s & 50.0s \\ 
        & DeepSeek-V3 & Text RAG & 48.5s & 7.5s & 56.0s & 48.5s & 21.4s & 69.9s \\ \midrule 

        \multirow{6}{*}{\textbf{\textit{LVLM-based}}} & \multirow{3}{*}{Qwen2.5-VL-3B} & Direct & - & 34.3s & 34.3s & - & 34.3s & 34.3s  \\ 
        & & M3DocRAG & 10.7s & 5.7s & 16.4s & 10.7s & 8.3s & 19.0s \\ 
        & & MoLoRAG & 38.4s & 5.7s & 44.1s & 38.4s & 8.2s & 46.6s \\ \cmidrule{2-9}
        &  \multirow{3}{*}{Qwen2.5-VL-7B} & Direct & - & 47.1s & 47.1s &  - & 47.1s & 47.1s \\ 
        & & M3DocRAG & 10.7s & 6.8s & 17.5s & 10.7s & 10.2s & 20.9s \\ 
        & & MoLoRAG & 38.4s & 6.8s & 45.2s & 38.4s & 10.1s & 48.5s \\ \midrule 
        \textbf{\textit{Multi-agent}} & \multicolumn{2}{c|}{MDocAgent (LLaMA3.1-8B+Qwen2.5-VL-7B)} & 31.8s & 16.8s & 48.6s  & 31.8s & 35.5s & 67.3s\\ 
      \bottomrule  
    \end{tabular}
    }
    \label{tab:total_time_comp}
\end{table*}

\begin{figure*}[!t]
    \centering
    \includegraphics[width=\linewidth]{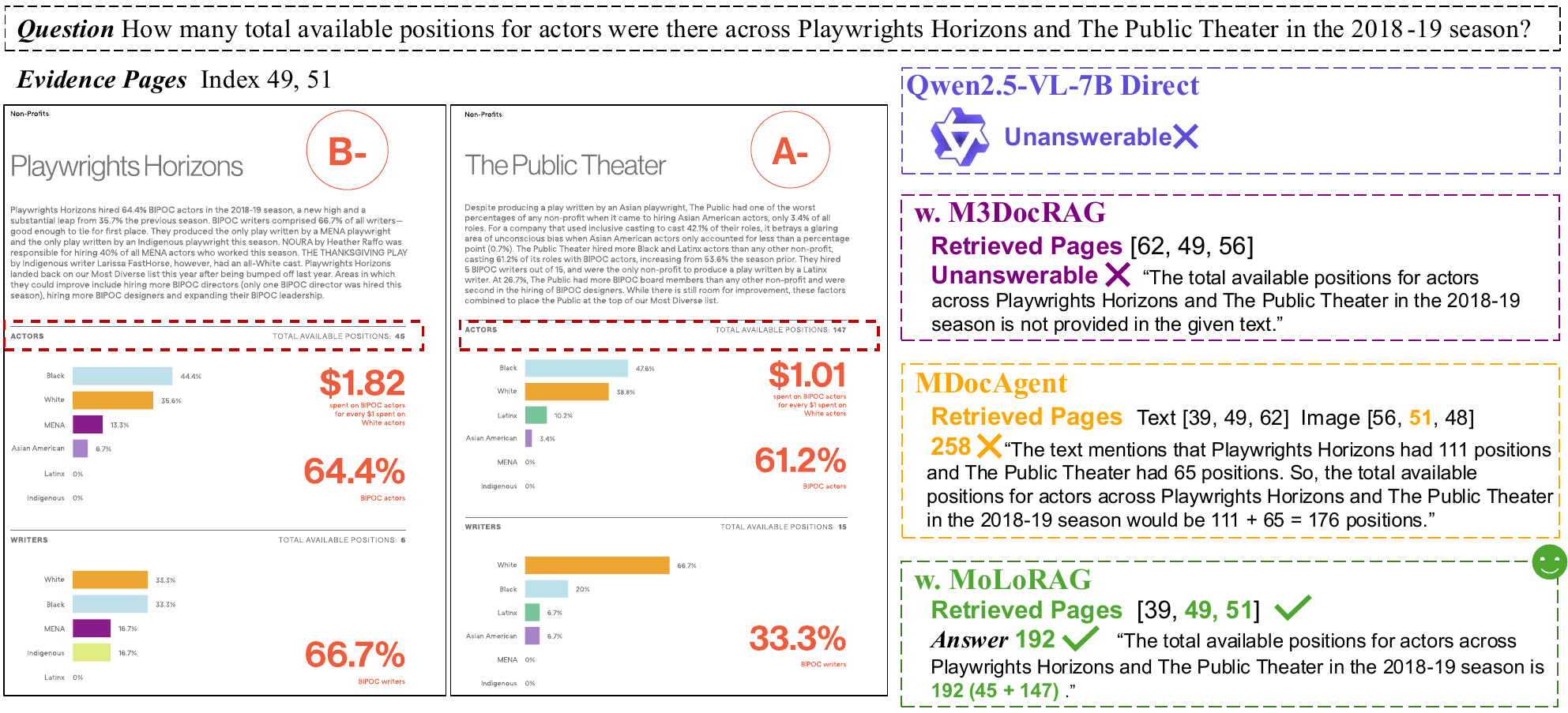}
    \caption{\textbf{Case study under the top-$3$ setting on LongDocURL involving cross-page reasoning.} The given question requires evidence from two separate pages to derive the correct answer. LVLM direct inference fails due to its limited context capacity (e.g., processing up to 30 pages). M3DocRAG is unable to retrieve the relevant content, leading to failure. MDocAgent relies on only one evidence page and falls into hallucination, producing an incorrect answer. In contrast, MoLoRAG successfully \textbf{retrieves both relevant pages}, enabling the LVLM to leverage this information to \textbf{answer the question correctly}.
    }
    \label{fig:case_study_2}
\end{figure*}

\noindent \textbf{Total Time Costs} In addition to inference time efficiency, we present the total time (retrieval and inference) for various methods to provide a clearer illustration. Note that the retrieval time includes both indexing and retrieval processes. DeepSeek-V3 and GPT-4o are invoked via APIs in a sequential manner to ensure a fair comparison, while the remaining open-source LLMs are deployed locally using vLLM on a single NVIDIA A6000-48G GPU. For LVLM Direct, the number of pages considered is up to 30, with times reported using 3 A6000 GPUs. All other time costs are measured on a single A6000 GPU, except for MDocAgent, which utilizes 2 A6000 GPUs. The results are shown in Table \ref{tab:total_time_comp}.

For LLM with Text RAG, variations in latency occur due to differences in invocation methods and model architecture. The index stage that invokes Qwen-RAG via APIs significantly affects overall latency. In LVLM-based methods, direct inference involves managing 30 pages, which can slow processing. While M3DocRAG is the most efficient method, it focuses solely on semantic relevance, limiting retrieval accuracy. The robust baseline, MDocAgent, employs parallel text and image retrieval; however, its multi-agent framework can hinder overall efficiency as context complexity increases. Despite the associated retrieval time costs, MoLoRAG maintains a notably fast inference stage and scales well with larger $K$ values, ensuring a balance between efficiency and performance.

\subsection{Case Study}

We present another case involving cross-page reasoning, as shown in Figure \ref{fig:case_study_2}. The question asks: ``How many total available positions for actors were there across Playwrights Horizons and The Public Theater in the 2018–19 season?'' Answering this requires evidence from two separate pages: the available positions for actors at Playwrights Horizons and The Public Theater.

LVLM direct inference fails due to limited context capacity, as it can only process up to 30 pages, while the evidence pages are located at indices 49 and 51. Both M3DocRAG and MDocAgent fail to retrieve these evidence pages because they rely solely on semantic relevance. Specifically, MDocAgent retrieves only one evidence page, resulting in hallucination and an incorrect answer. In contrast, MoLoRAG effectively retrieves both relevant pages, enabling the LVLM to access the necessary information and correctly compute the total available positions, demonstrating its superior cross-page reasoning capability.

\subsection{Ablation Study}\label{appendix:ablation} 

In this subsection, we present the ablation study to evaluate the effectiveness of individual components within MoLoRAG, focusing on two key aspects: the combination of semantic and logical relevance, and the graph construction process. 

\begin{table}[!t]
    \centering
    \caption{\textbf{Retrieval performance comparison of MoLoRAG\textsubscript{Logi} and MoLoRAG+ with the same fine-tuned retrieval engine.} MoLoRAG\textsubscript{Logi}, which relies solely on logical relevance, consistently underperforms compared to MoLoRAG+, demonstrating the effectiveness of combining semantic and logical relevance.}
    \resizebox{\linewidth}{!}{%
     \begin{tabular}{cc|cccc}
       \toprule
      \rowcolor{COLOR_MEAN} \textbf{Dataset} & \textbf{$K$} & \textbf{Method} & \textbf{Recall} & \textbf{NDCG}  & \textbf{MRR} \\ \midrule
       \multirow{4}{*}{\textbf{MMLong}}  & \multirow{2}{*}{$1$} & MoLoRAG\tiny{Logi} & 46.55 & 58.90 & 58.90 \\ 
       & & MoLoRAG+ & \textbf{51.32} & \textbf{66.86} & \textbf{66.86} \\ \cmidrule{3-6}

       & \multirow{2}{*}{$3$} & MoLoRAG\tiny{Logi} & 60.45 & 58.02 & 64.62 \\ 
       & & MoLoRAG+ & \textbf{68.87} & \textbf{64.49} & \textbf{73.50} \\ \midrule

      \multirow{4}{*}{\textbf{LongDocURL}} & \multirow{2}{*}{$1$} & MoLoRAG\tiny{Logi} & 47.65 & 65.56 & 65.56 \\ 
      & & MoLoRAG+ & \textbf{50.82} & \textbf{70.08} & \textbf{70.08} \\ \cmidrule{3-6} 

       & \multirow{2}{*}{$3$} & MoLoRAG\tiny{Logi} & 62.71 & 61.27 & 71.53 \\ 
       & & MoLoRAG+ & \textbf{68.92} & \textbf{64.90} & \textbf{77.14} \\ 
     \bottomrule
    \end{tabular}
    }
    \label{tab:ablation_logic}
    \vspace*{-10pt}
\end{table}

\noindent \textbf{Combination of Semantic and Logical Relevance} We consider a variant of MoLoRAG that relies solely on logical relevance by setting $s_i = s_i^{\text{logi}}$ within the framework. This variant is referred to as MoLoRAG$_\text{logi}$. In this setup, the retrieval engine is the fine-tuned Qwen2.5-VL-3B, which already demonstrates strong task understanding. We compare the retrieval performance of MoLoRAG$_\text{logi}$ with MoLoRAG+ in Table \ref{tab:ablation_logic}. From the results, it becomes evident that relying solely on logical relevance does not achieve optimal performance. This is because VLMs may exhibit over-confidence and, in some cases, fall into hallucinations when relying exclusively on reasoning capabilities. Additionally, logical relevance scores are discrete, often leading to multiple pages with identical scores, making it difficult to rank and distinguish between them. Consequently, it is more reliable to combine both logical and semantic relevance.

\noindent \textbf{Effectiveness of Graph Construction} We evaluate another variant of MoLoRAG, named MoLoRAG\textsubscript{Full}, in which the VLM \textbf{traverses all pages within the document} for a given question, assigning a relevance score to each page. Each page's relevance score is updated by combining semantic and logical relevance scores. After traversal, only the top-$K$ pages are retained as the final retrieval result. Unlike MoLoRAG, this variant \textbf{eliminates the graph-based indexing mechanism} and instead \textbf{sequentially processes all pages in the document for selection}. We compare the retrieval performance of this variant with MoLoRAG+ in Table \ref{tab:ablation_fulldoc}. While MoLoRAG\textsubscript{Full} achieves a slight improvement due to the expanded set of queried pages, it requires nearly twice the computational time (Figure \ref{fig:retrieval_scalability}), significantly reducing efficiency. Furthermore, this marginal performance difference highlights the effectiveness of our graph construction, which provides a high-quality candidate set for retrieval.

\begin{table}[!t]
    \centering
    \caption{\textbf{Retrieval performance comparison between MoLoRAG\textsubscript{Full} and MoLoRAG+ using the same fine-tuned retrieval engine.} MoLoRAG\textsubscript{Full} sequentially queries each page of the entire document and combines their relevance scores for final re-ranking. Although this variant achieves a marginal improvement over MoLoRAG+, it requires nearly \textbf{twice the computational time} (Figure \ref{fig:retrieval_scalability}), significantly reducing efficiency.}
    \resizebox{\linewidth}{!}{%
     \begin{tabular}{cc|cccc}
       \toprule
      \rowcolor{COLOR_MEAN} \textbf{Dataset} & \textbf{$K$} & \textbf{Method} & \textbf{Recall} & \textbf{NDCG}  & \textbf{MRR} \\ \midrule
       \multirow{4}{*}{\textbf{MMLong}}  & \multirow{2}{*}{$1$} & MoLoRAG\tiny{Full} &  \textbf{51.63} & \textbf{67.21} & \textbf{67.21} \\ 
       & & MoLoRAG+ & 51.32 & 66.86 & 66.86 \\ \cmidrule{3-6}

       & \multirow{2}{*}{$3$} & MoLoRAG\tiny{Full} & \textbf{73.64} & 64.31 & \textbf{75.20} \\ 
       & & MoLoRAG+ & 68.87 & \textbf{64.49} & 73.50 \\ \midrule

      \multirow{4}{*}{\textbf{LongDocURL}} & \multirow{2}{*}{$1$} & MoLoRAG\tiny{Full} &  \textbf{51.24} & \textbf{70.68} & \textbf{70.68} \\ 
      & & MoLoRAG+ & 50.82 & 70.08 & 70.08 \\ \cmidrule{3-6} 

       & \multirow{2}{*}{$3$} & MoLoRAG\tiny{Full} & \textbf{72.30} & 64.32 & \textbf{78.53}  \\ 
       & & MoLoRAG+ & 68.92 & \textbf{64.90} & 77.14 \\ 
     \bottomrule
    \end{tabular}
    }
    \label{tab:ablation_fulldoc}
    \vspace*{-10pt}
\end{table}

\subsection{Fine-grained Analysis}\label{appendix:fine_grained}
This subsection provides a detailed performance analysis across evidence modalities (e.g., text, tables, figures) and question contexts (e.g., single-page and multi-page understanding). The analysis focuses on the MMLongBench and LongDocURL datasets, which offer official splits based on modalities and locations. For MMLongBench, results for top-$1$, top-$3$, and top-$5$ settings are presented in Tables \ref{tab:fine_grained_top1_mmlong}, \ref{tab:fine_grained_top3_mmlong}, and \ref{tab:fine_grained_top5_mmlong}, respectively. For LongDocURL, the corresponding results are shown in Tables \ref{tab:fine_grained_longdoc}, \ref{tab:fine_grained_longdoc_top3}, and \ref{tab:fine_grained_longdoc_top5}. From these results, we conclude that \textbf{LVLMs w. MoLoRAG excel across diverse modalities}. While LLMs w. Text RAG methods perform reasonably well on text-based modalities, they struggle significantly with non-textual content like figures (e.g., 9.83\% on Figure). This highlights the critical need for robust multi-modal understanding. In contrast, LVLMs integrated with MoLoRAG demonstrate strong and balanced performance across all modalities. For example, Qwen2.5-VL-7B with MoLoRAG+ achieves 37.43\% on Text and 36.94\% on Figure (MMLongBench in the top-1 setting), showing the effectiveness of retrieval-based multi-modal reasoning.

\input{tables/main_top1}
\input{tables/main_top5}

\input{tables/detail_mmlong}
\input{tables/detail_longdocurl}

%% file: tables/main_top1.tex
\begin{table*}[!t]
    \centering
    \caption{\textbf{Overall performance comparison (in $\%$) under the retrieved top-$1$ setting.} The best performance across all methods is \colorbox{orange!20}{\textbf{highlighted}}. }
    \vspace*{-8pt}
    \resizebox{0.95\linewidth}{!}{

        \begin{tabular}{ccc|cccc|c}
      \toprule
      \rowcolor{COLOR_MEAN} \textbf{Type} & \textbf{Model} & \textbf{Method}  & \textbf{MMLongBench}  & \textbf{LongDocURL} & \textbf{PaperTab} & \textbf{FetaTab} & \textbf{Avg.} \\ \midrule 
      \multirow{5}{*}{\textit{\textbf{LLM-based}}} &
      Mistral-7B & Text RAG & 21.66 & 17.63 & 5.09 & 24.11 & 17.12  \\ 
      & Qwen2.5-7B & Text RAG & 22.11 & 20.75 & 5.34 & 22.64 & 17.71 \\ 
     & LLaMA3.1-8B & Text RAG & 18.70 & 22.57 & 7.12 & 28.25 & 19.16 \\ 
     & GPT-4o & Text RAG & 24.07 & 22.84 & 8.65 & 34.15 & 22.43 \\ 
    &  DeepSeek-v3 & Text RAG & \textbf{25.94} & \textbf{24.32} & \textbf{10.18} & \textbf{34.55} & \textbf{23.75} \\  \midrule

     \multirow{16}{*}{\textit{\textbf{LVLM-based}}} & \multirow{4}{*}{LLaVA-Next-7B} & Direct & 7.15 & 10.78 & 3.05 & 11.61 & 8.15 \\ 
     & & M3DocRAG & 16.32 & 25.25 & 6.62 & 15.26 & 15.86 \\ 
     & & MoLoRAG & 16.73 & 26.11 & \textbf{6.87} & \textbf{18.41} & \textbf{17.03} \\ 
     & & MoLoRAG+ & \textbf{17.15} & \textbf{27.00} & 6.36 & 17.52 & 17.01 \\ \cmidrule{2-8}

      & \multirow{4}{*}{DeepSeek-VL-16B} & Direct & 8.40 & 14.72 & 6.11 & 16.14 & 11.34 \\ 
     &  & M3DocRAG & 26.23 & 42.21 & 16.54 & 48.43 & 33.35 \\ 
     & & MoLoRAG & 27.47 & 44.75 & 20.87 & 56.89 & 37.50  \\ 
     & & MoLoRAG+ & \textbf{28.98} & \textbf{45.17} & \textbf{21.88} & \textbf{58.27} & \textbf{38.58} \\ \cmidrule{2-8}
     
      &  \multirow{4}{*}{Qwen2.5-VL-3B} & Direct & 26.65 & 24.89 & \textbf{25.19} & 51.57 & 32.08 \\ 
     &  & M3DocRAG & 26.77 & 39.82 & 19.85 & 45.77 & 33.05 \\ 
     & & MoLoRAG & 29.08 & 41.95 & 21.88 & 54.72 & 36.91 \\ 
     & & MoLoRAG+ & \textbf{30.03} & \textbf{43.17} & 23.16 & \textbf{55.41} & \textbf{37.94} \\ \cmidrule{2-8}
        
     & \multirow{4}{*}{Qwen2.5-VL-7B} & Direct & 32.77 & 26.38 &  \cellcolor{orange!20} \textbf{29.77} & \cellcolor{orange!20} \textbf{64.07} & 38.25 \\ 
      & & M3DocRAG & 32.29  & 43.32 & 19.34 & 50.98 & 36.48 \\ 
 
     &  & MoLoRAG & 34.35 & 46.89 & 23.92 & 61.52 & 41.67 \\ 
      & & MoLoRAG+ & \cellcolor{orange!20} \textbf{36.37} & \cellcolor{orange!20}  \textbf{47.86} & 27.48 & 62.50 & \cellcolor{orange!20}  \textbf{43.55} \\ \midrule 
     \textbf{\textit{Multi-agent}} & \multicolumn{2}{c|}{MDocAgent (LLaMA3.1-8B + Qwen2.5-VL-7B)} & 31.73 & 44.42 & 21.63 & 57.78 & 38.89 \\ 
      \bottomrule
    \end{tabular}
    
    }
    \label{tab:main_top1}
\end{table*}

%% file: tables/main_top5.tex
\begin{table*}[!t]
    \centering
    \caption{\textbf{Overall performance comparison (in $\%$) under the retrieved top-$5$ setting.} The best performance across all methods is \colorbox{orange!20}{\textbf{highlighted}}. }
    \vspace*{-8pt}
    \resizebox{0.95\linewidth}{!}{

    \begin{tabular}{ccc|cccc|c}
      \toprule
      \rowcolor{COLOR_MEAN} \textbf{Type} & \textbf{Model} & \textbf{Method}  & \textbf{MMLongBench}  & \textbf{LongDocURL} & \textbf{PaperTab} & \textbf{FetaTab} & \textbf{Avg.} \\ \midrule 
     \multirow{5}{*}{\textit{\textbf{LLM-based}}} &  Mistral-7B & Text RAG & 23.43 & 26.43 & 13.23 & 48.62 & 27.93 \\ 
     & Qwen2.5-7B & Text RAG & 26.09 & 31.36 & 16.79 & 49.21 & 30.86 \\ 
    & LLaMA3.1-8B & Text RAG & 24.25 & 33.27 & 17.81 & 54.53 & 32.47  \\ 
    & GPT-4o & Text RAG & 28.74 & 36.98 & 20.36 & 57.78 & 35.97  \\ 
     & DeepSeek-v3 & Text RAG & \textbf{31.23} & \textbf{39.04} & \textbf{23.92} & \textbf{62.01} & \textbf{39.05} \\  \midrule

    \multirow{16}{*}{\textbf{\textit{LVLM-based}}} &  \multirow{4}{*}{LLaVA-Next-7B} & Direct & 7.15 & 10.78 & 3.05 & 11.61 & 8.15 \\ 
     & & M3DocRAG & \textbf{10.43} & 12.65 & \textbf{4.58} & 12.80 & \textbf{10.12}  \\ 
     & & MoLoRAG & 9.56 & 12.72 & 4.07 & \textbf{14.07} & 10.11 \\ 
     & & MoLoRAG+ & 9.19 & \textbf{13.59} & 4.33 & 13.09 & 10.05 \\ \cmidrule{2-8}

     & \multirow{4}{*}{DeepSeek-VL-16B} & Direct & 8.40 & 14.72 & 6.11 & 16.14 & 11.34 \\ 
     & & M3DocRAG & 18.87 & 29.27 & 8.14 & 27.26 & 20.89 \\ 
     & & MoLoRAG & 20.07 & 30.76 & 8.40 & 39.76 & 24.75 \\ 
     & & MoLoRAG+ & \textbf{24.86} & \textbf{38.02} & \textbf{9.67} & \textbf{41.44} & \textbf{28.50} \\ \cmidrule{2-8}
     
      &   \multirow{4}{*}{Qwen2.5-VL-3B} & Direct & 26.65 & 24.89 & 25.19 & 51.57 & 32.08 \\ 
     & & M3DocRAG & 28.38 & 44.67 & 27.48 & 55.22 & 38.94 \\
      & & MoLoRAG & 31.43 & \textbf{46.05} & 26.97 & 57.48 & 40.48 \\ 
      & & MoLoRAG+ & \textbf{32.41} & 45.13 & \textbf{27.48} & \textbf{58.07} & \textbf{40.77} \\ \cmidrule{2-8}
        
     & \multirow{4}{*}{Qwen2.5-VL-7B} & Direct & 32.77 & 26.38 & 29.77 & 64.07 & 38.25 \\ 
     & & M3DocRAG & 37.19 & 50.33 & 30.53 & 64.37 & 45.61 \\ 
     & & MoLoRAG & 39.97 & 52.33 & 31.04 & 68.80 & 48.04  \\ 
     & & MoLoRAG+ & \cellcolor{orange!20} \textbf{40.47} &  \cellcolor{orange!20}  \textbf{52.33} & \cellcolor{orange!20}  \textbf{31.55} & \cellcolor{orange!20}  \textbf{69.39} & \cellcolor{orange!20}  \textbf{48.44}  \\ \midrule

     \textbf{\textit{Multi-agent}} & \multicolumn{2}{c|}{MDocAgent (LLaMA3.1-8B + Qwen2.5-VL-7B)} & 38.34 & 48.07 & 29.77 & 63.78 & 44.99 \\ 
       
      \bottomrule
    \end{tabular}

    }
    \label{tab:main_top5}
\end{table*}

%% file: tables/detail_mmlong.tex
\begin{table*}[!t]
    \centering
     \caption{\textbf{Fine-grained performance analysis (Accuracy in $\%$) across evidence modality and evidence locations on MMLongBench under the retrieved top-$1$ setting.} ``UNA'' refers to unanswerable questions.}
     \vspace*{-8pt}
    \resizebox{\linewidth}{!}{

    \begin{tabular}{ccc|ccccc|ccc|cc}
      \toprule
      \rowcolor{COLOR_MEAN} & &  & \multicolumn{5}{c|}{\textbf{Modality}} & \multicolumn{3}{c|}{\textbf{Location}} & \multicolumn{2}{c}{\textbf{Overall}} \\ 
      \rowcolor{COLOR_MEAN} \multirow{-2}{*}{\textbf{Type}} &  \multirow{-2}{*}{\textbf{Model}} & \multirow{-2}{*}{\textbf{Method}} & Text & Table & Figure & Chart & Layout & Single & Multiple & UNA & Acc & EM \\ \midrule
     \multirow{5}{*}{\textbf{\textit{LLM-based}}} &  Mistral-7B & Text RAG & 11.32 & 6.60 & 5.20 & 6.97 & 7.21 & 10.54 & 4.26 & 75.78 & 21.66 & 19.96  \\ 
     & Qwen2.5-7B & Text RAG & 13.31 & 9.56 & 6.65 & 6.18 & 8.01 & 12.09 & 5.41 & 73.09 & 22.11 & 20.61 \\ 
     & LLaMA3.1-8B & Text RAG & 16.53 & 10.93 & 7.80 & 9.84 & \textbf{12.92} & 13.26 & 7.96 & 47.98 & 18.70 & 16.36 \\ 
     & GPT-4o & Text RAG & 15.04 & 12.95 & 6.39 & 8.05 & 10.33 & 13.34 & 5.97 & \textbf{77.13} & 24.07 & 22.46 \\ 
     & DeepSeek-v3 & Text RAG & \textbf{17.76} & \textbf{14.70} & \textbf{9.83} & \textbf{9.99} & 10.72 & \textbf{16.80} & \textbf{7.97} & 76.68 & \textbf{25.94} & \textbf{23.84}  \\  \midrule 

     \multirow{16}{*}{\textbf{\textit{LVLM-based}}} &  \multirow{4}{*}{LLaVA-Next-7B} & Direct & 6.54 & 1.53 & 7.38 & 2.12 & 4.38 & 4.17 & 5.07 & \textbf{16.59} & 7.15 & 5.73 \\ 
     & & M3DocRAG & 18.04 & \textbf{10.47} & 20.03 & 13.35 & 14.76 & 19.93 & 11.26 & 15.70 & 16.32 & 13.03 \\ 
     & & MoLoRAG & 18.12 & 9.71 & 19.68 & \textbf{13.57} & \textbf{16.57} & 19.80 & \textbf{12.13} & 16.14 & 16.73 & 12.94 \\ 
     & & MoLoRAG+ & \textbf{18.94} & 9.85 & \textbf{21.61} & 13.22 & 15.94 & \textbf{22.92} & 10.50 & 13.90 & \textbf{17.15} & \textbf{13.03} \\ \cmidrule{2-13}

      & \multirow{4}{*}{DeepSeek-VL-16B} & Direct & 8.86 & 6.57 & 13.39 & 5.23 & 13.63 & 9.86 & 9.36 & 3.14 & 8.40 & 6.01 \\ 
     & & M3DocRAG & 30.84 & 23.95 & 31.55 & 27.29 & 30.16 & 41.55 & 15.76 & 7.62 & 26.23 & 20.98 \\ 
     & & MoLoRAG & 31.67 & \textbf{28.54} & 30.80 & 26.42 & \textbf{32.84} & 43.65 & 16.63 & \textbf{8.07} & 27.47 & 21.81 \\ 
     & & MoLoRAG+ & \textbf{35.14} & 27.86 & \textbf{35.32} & \textbf{27.42} & 30.74 & \textbf{47.62} & \textbf{17.37} & 4.93 & \textbf{28.98} & \textbf{22.83} \\ \cmidrule{2-13}
     
      &  \multirow{4}{*}{Qwen2.5-VL-3B} & Direct & 34.11 & 23.37 & \textbf{33.75} & 24.72 & \textbf{29.56} & 36.30 & \textbf{23.42} & 9.87 & 26.65 & 20.98 \\ 
     & & M3DocRAG & 31.88 & 22.70 & 27.87 & 23.85 & 22.22 & 37.90 & 15.40 & 20.63 & 26.77 & 21.90 \\ 
     & & MoLoRAG & 33.07 & \textbf{29.18} & 29.06 & 23.44 & 24.88 & 41.58 & 17.02 & \textbf{21.52} & 29.08 & 23.84  \\ 
     & & MoLoRAG+ & \textbf{35.26} & 28.11 & 32.09 & \textbf{25.03} & 26.66 & \textbf{43.68} & 18.11 & 19.73 & \textbf{30.03} & \textbf{24.49} \\ \cmidrule{2-13}

     & \multirow{4}{*}{Qwen2.5-VL-7B} & Direct & 37.14 & 25.52 & 31.95 & 28.00 & 27.26 & 40.21 & \textbf{23.88} & 30.94 & 32.77 & 27.36 \\ 
     & & M3DocRAG & 31.87 & 23.88 & 30.03 & 26.48 & 26.74 & 42.66 & 12.37 & \textbf{40.81} & 32.29 & 27.63 \\ 
     & & MoLoRAG & 33.80 & \textbf{32.12} & 29.77 & 29.23 & 30.34 & 46.61 & 15.37 & 37.22 & 34.35 & 29.67 \\ 
     & & MoLoRAG+ & \textbf{37.43} & 31.34 & \textbf{36.94} & \textbf{29.95} & \textbf{33.11} & \textbf{50.14} & 18.25 & 34.08 & \textbf{36.37} & \textbf{30.87} \\ \midrule

     \textbf{\textit{Multi-agent}} & \multicolumn{2}{c|}{MDocAgent (LLaMA3.1-8B+Qwen2.5-VL-7B)} & 35.08 & 30.13 & 29.61 & 27.47 & 24.72 & 45.26 & 14.86 & 29.15 & 31.73 & 27.45 \\ 
       
      \bottomrule
    \end{tabular}
    
    }
    \label{tab:fine_grained_top1_mmlong}
\end{table*}

\begin{table*}[!t]
    \centering
     \caption{\textbf{Fine-grained performance analysis (Accuracy in $\%$) across evidence modality and evidence locations on MMLongBench under the retrieved top-$3$ setting.} ``UNA'' refers to unanswerable questions. }
     \vspace*{-8pt}
    \resizebox{\linewidth}{!}{
    \begin{tabular}{ccc|ccccc|ccc|cc}
      \toprule
      \rowcolor{COLOR_MEAN} &  &  & \multicolumn{5}{c|}{\textbf{Modality}} & \multicolumn{3}{c|}{\textbf{Location}} & \multicolumn{2}{c}{\textbf{Overall}} \\ 
      \rowcolor{COLOR_MEAN} \multirow{-2}{*}{\textbf{Type}} &  \multirow{-2}{*}{\textbf{Model}} & \multirow{-2}{*}{\textbf{Method}} & Text & Table & Figure & Chart & Layout & Single & Multiple & UNA & Acc & EM \\ \midrule
     \multirow{5}{*}{\textbf{\textit{LLM-based}}} &  Mistral-7B & Text RAG &  15.97 & 14.77 & 8.41 & 10.70 & 12.38 & 16.16 & 8.86 & 68.61 & 24.47 & 22.00 \\ 
     & Qwen2.5-7B & Text RAG & 17.96 & 15.42 & 9.58 & 10.35 & 10.69 & 17.72 & 9.34 & 70.40 & 25.52 & 23.29 \\ 
     & LLaMA3.1-8B & Text RAG & 20.40 & 20.02 & 10.44 & 15.42 & 13.82 & 18.74 & \textbf{13.62} & 45.29 & 22.56 & 19.22 \\ 
     & GPT-4o & Text RAG & 19.68 & 19.14 & 10.10 & 13.58 & 12.25 & 20.20 & 10.52 & \textbf{70.40} & 27.23 & 24.31 \\ 
     & DeepSeek-v3 & Text RAG & \textbf{25.37} & \textbf{22.23} & \textbf{13.34} & \textbf{19.60} & \textbf{17.27} & \textbf{24.85} & 13.03 & 69.06 & \textbf{29.82} & \textbf{26.62}  \\  \midrule 

    \multirow{16}{*}{\textbf{\textit{LVLM-based}}} &  \multirow{4}{*}{LLaVA-Next-7B} &  Direct & 6.54 & 1.53 & 7.38 & 2.12 & 4.38 & 4.17 & 5.07 & 16.59 & 7.15 & 5.73 \\ 
     & & M3DocRAG & \textbf{8.74} & \textbf{6.59} & \textbf{11.72} & 1.87 & \textbf{8.54} & 7.27 & \textbf{8.55} & \textbf{17.49} & \textbf{10.10} & \textbf{8.23} \\ 
     & & MoLoRAG & 6.84 & 5.55 & 10.72 & 2.15 & 7.66 & 7.82 & 6.43 & 16.14 & 9.37 & 7.49 \\ 
     & & MoLoRAG+ & 7.49 & 2.49 & 11.24 & \textbf{2.89} & 8.08 & \textbf{7.86} & 6.25 & 16.59 & 9.41 & 7.30 \\ \cmidrule{2-13}

     & \multirow{4}{*}{DeepSeek-VL-16B} & Direct & 8.86 & 6.57 & 13.39 & 5.23 & 13.63 & 9.86 & 9.36 & 3.14 & 8.40 & 6.01 \\ 
     & & M3DocRAG & 19.75 & 14.31 & 27.55 & 18.38 & 25.02 & 25.91 & 15.84 & 3.59 & 18.12 & 13.49 \\ 
     & & MoLoRAG & 21.57 & 18.79 & 29.00 & 17.55 & 24.54 & 29.60 & 18.15 & 2.69 & 20.43 & 16.27 \\ 
     & & MoLoRAG+ & \textbf{27.58} & \textbf{23.33} & \textbf{34.45} & \textbf{21.56} & \textbf{32.67} & \textbf{39.40} & \textbf{18.74} & \textbf{4.04} & \textbf{25.47} & \textbf{19.59}  \\ \cmidrule{2-13}
     
       & \multirow{4}{*}{Qwen2.5-VL-3B} & Direct & 34.11 & 23.37 & 33.75 & 24.72 & 29.56 & 36.30 & 23.42 & 9.87 & 26.65 & 20.98 \\ 
     & & M3DocRAG & 35.11 & 25.58 & 32.23 & 24.04 & 28.06 & 39.62 & 20.62 & 18.83 & 29.11 & 23.66 \\ 
     & & MoLoRAG & \textbf{38.94} & \textbf{30.48} & \textbf{36.05} & 24.33 & 28.73 & 44.29 & \textbf{23.52} & 19.28 & 32.11 & 26.16 \\ 
     & & MoLoRAG+ & 38.29 & 29.39 & 35.34 & \textbf{26.48} & \textbf{33.53} & \textbf{45.12} & 23.17 & \textbf{19.28} & \textbf{32.47} & \textbf{26.52} \\ \cmidrule{2-13}
        
     & \multirow{4}{*}{Qwen2.5-VL-7B} & Direct & 37.14 & 25.52 & 31.95 & 28.00 & 27.26 & 40.21 & 23.88 & 30.94 & 32.77 & 27.36 \\ 
     & & M3DocRAG &  38.83 & 36.24 & 35.83 & 30.46 & 36.56 & 46.85 & 25.29 & 28.70 & 36.18 & 30.96 \\ 
     & & MoLoRAG & 41.67 & 37.89 & 37.56 & \textbf{34.15} & 32.44 & 50.07 & 26.12 & 34.98 & 39.28 & 33.18 \\ 
     & & MoLoRAG+ & \textbf{42.69} & \textbf{38.53} & \textbf{40.73} & 33.26 & \textbf{38.79} & \textbf{52.90} & \textbf{27.59} & \textbf{35.87} & \textbf{41.01} & \textbf{34.94} \\  \midrule

     \textbf{\textit{Multi-agent}} & \multicolumn{2}{c|}{MDocAgent (LLaMA3.1-8B+Qwen2.5-VL-7B)} & 43.14 & 38.72 & 37.90 & 32.55 & 31.17 & 53.45 & 23.82 & 28.25 & 38.53 & 33.27 \\ 
       
      \bottomrule
    \end{tabular}
    
    }
    \label{tab:fine_grained_top3_mmlong}
\end{table*}

\begin{table*}[!t]
    \centering
     \caption{\textbf{Fine-grained performance analysis (Accuracy in $\%$) across evidence modality and evidence locations on MMLongBench under the retrieved top-$5$ setting.} ``UNA'' refers to unanswerable questions.}
     \vspace*{-8pt}
    \resizebox{\linewidth}{!}{

    \begin{tabular}{ccc|ccccc|ccc|cc}
      \toprule
      \rowcolor{COLOR_MEAN} & &  & \multicolumn{5}{c|}{\textbf{Modality}} & \multicolumn{3}{c|}{\textbf{Location}} & \multicolumn{2}{c}{\textbf{Overall}} \\ 
      \rowcolor{COLOR_MEAN} \multirow{-2}{*}{\textbf{Type}} & \multirow{-2}{*}{\textbf{Model}} & \multirow{-2}{*}{\textbf{Method}} & Text & Table & Figure & Chart & Layout & Single & Multiple & UNA & Acc & EM \\ \midrule
     \multirow{4}{*}{ \textbf{\textit{LLM-based}}} & Mistral-7B & Text RAG & 17.41 & 12.03 & 8.13 & 13.32 & 16.30 & 16.16 & 10.59 & 61.43 & 23.43 & 20.43  \\ 
     & Qwen2.5-7B & Text RAG & 19.98 & 19.29 & 10.06 & 12.57 & 15.31 & 19.99 & 11.34 & 64.13 & 26.09 & 23.57 \\ 
     & LLaMA3.1-8B & Text RAG & 24.61 & 22.71 & 12.21 & 18.42 & \textbf{21.49} & 22.60 & \textbf{15.74} & 41.26 & 24.25 & 21.07 \\ 
     & GPT-4o & Text RAG & 22.38 & 24.50 & 12.30 & 15.42 & 16.17 & 23.37 & 13.54 & \textbf{65.47} & 28.74 & 25.51 \\ 
     & DeepSeek-v3 & Text RAG & \textbf{27.54} & \textbf{28.33} & \textbf{15.53} & \textbf{21.39} & 20.44 & \textbf{28.90} & 15.67 & 62.33 & \textbf{31.23} & \textbf{27.54}  \\  \midrule 

     \multirow{16}{*}{\textbf{\textit{LVLM-based}}} & \multirow{4}{*}{LLaVA-Next-7B} & Direct & 6.54 & 1.53 & 7.38 & 2.12 & 4.38 & 4.17 & 5.07 & 16.59 & 7.15 & 5.73 \\ 
     & & M3DocRAG & \textbf{8.59} & \textbf{6.23} & \textbf{12.16} & 3.40 & 7.43 & \textbf{8.65} & \textbf{7.92} & \textbf{17.49} & \textbf{10.43} & \textbf{7.95} \\ 
     & & MoLoRAG & 7.33 & 5.52 & 11.36 & 3.12 & 8.34 & 7.95 & 7.12 & 16.14 & 9.56 & 7.67 \\ 
     & & MoLoRAG+ & 6.46 & 2.26 & 10.33 & \textbf{3.70} & \textbf{8.54} & 8.02 & 4.79 & 17.49 & 9.19 & 7.30 \\ \cmidrule{2-13} 

     & \multirow{4}{*}{DeepSeek-VL-16B} & Direct & 8.86 & 6.57 & 13.39 & 5.23 & 13.63 & 9.86 & 9.36 & 3.14 & 8.40 & 6.01 \\
     & & M3DocRAG & 22.04 & 15.45 & 28.86 & 15.25 & 23.28 & 26.90 & 15.77 & \textbf{4.93} & 18.87 & 14.51 \\ 
     & & MoLoRAG & 22.38 & 18.64 & 28.07 & 13.83 & 23.68 & 29.60 & 16.52 & 4.04 & 20.07 & 15.80 \\ 
     & & MoLoRAG+ & \textbf{26.61} & \textbf{22.98} & \textbf{34.82} & \textbf{19.32} & \textbf{32.11} & \textbf{38.27} & \textbf{18.58} & 4.04 & \textbf{24.86} & \textbf{19.41}  \\ \cmidrule{2-13}
     
      &  \multirow{4}{*}{Qwen2.5-VL-3B} & Direct & 34.11 & 23.37 & 33.75 & 24.72 & 29.56 & 36.30 & 23.42 & 9.87 & 26.65 & 20.98 \\ 
     & & M3DocRAG & 35.79 & 26.04 & 32.06 & 24.15 & 29.33 & 39.08 & 22.16 & 14.35 & 28.38 & 22.46 \\ 
     & & MoLoRAG & 38.22 & \textbf{31.23} & 32.84 & \textbf{27.85} & 30.33 & 43.33 & 23.60 & 17.04 & 31.43 & 25.60 \\ 
     & & MoLoRAG+ & \textbf{38.38} & 30.99 & \textbf{36.04} & 26.00 & \textbf{33.94} & \textbf{44.82} & \textbf{23.67} & \textbf{18.83} & \textbf{32.41} & \textbf{26.52} \\ \cmidrule{2-13}

     & \multirow{4}{*}{Qwen2.5-VL-7B} & Direct & 37.14 & 25.52 & 31.95 & 28.00 & 27.26 & 40.21 & 23.88 & 30.94 & 32.77 & 27.36 \\  
     &  & M3DocRAG & 40.17 & 34.20 & 36.39 & \textbf{35.34} & 31.71 & 48.36 & 24.59 & 31.39 & 37.19 & 32.16 \\ 
     & & MoLoRAG & \textbf{43.07} & 38.10 & 38.92 & 35.22 & 35.08 & \textbf{51.72} & \textbf{27.02} & 33.63 & 39.97 & 34.20 \\ 
     & & MoLoRAG+ & 41.57 & \textbf{38.31} & \textbf{39.08} & 31.64 & \textbf{38.62} & 51.16 & 26.96 & \textbf{38.57} & \textbf{40.47} & \textbf{34.57} \\ \midrule 

     \textbf{\textit{Multi-agent}} & \multicolumn{2}{c|}{MDocAgent (LLaMA3.1-8B+Qwen2.5-VL-7B)}  & 41.92 & 42.00 & 34.02 & 33.45 & 29.77 & 49.97 & 25.25 & 32.29 & 38.34 & 32.99 \\ 
       
      \bottomrule
    \end{tabular}
    
    }
    \label{tab:fine_grained_top5_mmlong}
\end{table*}

%% file: tables/detail_longdocurl.tex
\begin{table*}[!t]
    \centering
    \caption{\textbf{Fine-grained performance analysis (Accuracy in $\%$) across evidence modality and evidence locations on LongDocURL under the retrieved top-$1$ setting.}}
    \vspace*{-8pt}
    \resizebox{\linewidth}{!}{

    \begin{tabular}{ccc|cccc|cc|cc}
      \toprule
      \rowcolor{COLOR_MEAN} & &  & \multicolumn{4}{c|}{\textbf{Modality}} & \multicolumn{2}{c|}{\textbf{Location}} & \multicolumn{2}{c}{\textbf{Overall}} \\ 
      \rowcolor{COLOR_MEAN} \multirow{-2}{*}{\textbf{Type}} &  \multirow{-2}{*}{\textbf{Model}} & \multirow{-2}{*}{\textbf{Method}} & Text & Table & Figure &  Layout & Single & Multiple  & Acc & EM \\ \midrule
      \multirow{5}{*}{\textbf{\textit{LLM-based}}} & Mistral-7B & Text RAG & 26.18 & 11.51 & 14.59 & 12.62 & 16.32 & 18.49 & 17.63 & 15.05 \\ 
     & Qwen2.5-7B & Text RAG & 29.25 & 14.98 & 20.50 & 16.10 & 20.66 & 20.53 & 20.75 & 17.46 \\ 
     & LLaMA3.1-8B & Text RAG & 30.63 & 16.26 & 22.18 & 16.52 & 21.32 & 23.49 & 22.57 & 18.11 \\ 
     & GPT-4o & Text RAG & 32.16 & 16.35 & 23.21 & 17.44 & 21.91 & 23.40 & 22.84 & 18.45 \\ 
     & DeepSeek-V3 & Text RAG & \textbf{33.28} & \textbf{18.17} & \textbf{25.47} & \textbf{19.84} & \textbf{24.00} & \textbf{24.34} & \textbf{24.32} & \textbf{20.09} \\  \midrule 

     \multirow{16}{*}{\textbf{\textit{LVLM-based}}} & \multirow{4}{*}{LLaVA-Next-7B} & Direct & 16.79 & 5.28 & 12.12 & 7.39 & 7.87 & 13.43 & 10.78 & 9.29   \\ 
     &  & M3DocRAG & 33.67 & 17.46 & 31.41 & 22.21 & 24.72 & 25.58 & 25.25 & 17.33  \\ 
     &  & MoLoRAG &  34.61 & 18.32 & 32.37 & 24.13 & 24.99 & \textbf{26.99} & 26.11 & 17.89 \\ 
     &  & MoLoRAG+ & \textbf{34.87} & \textbf{19.99} & \textbf{32.83} & \textbf{24.47} & \textbf{27.09} & 26.80 & \textbf{27.00} & \textbf{18.28}  \\ \cmidrule{2-11} 

     & \multirow{4}{*}{DeepSeek-VL-16B} & Direct & 19.98 & 8.26 & 13.81 & 13.65 & 11.18 & 17.87 & 14.72 & 11.35  \\ 
     & & M3DocRAG & 51.63 & 36.79 & 40.39 & 33.04 & 46.83 & 38.10 & 42.21 & 33.08 \\ 
     & & MoLoRAG & 54.64 & 39.97 & 41.52 & 34.64 & 48.77 & \textbf{41.17} & 44.75 & 35.18  \\ 
     & & MoLoRAG+ & \textbf{54.91} & \textbf{40.11} & \textbf{42.90} & \textbf{34.90} & \textbf{49.88} & 41.16 & \textbf{45.17} & \textbf{35.61} \\ \cmidrule{2-11} 
     
       & \multirow{4}{*}{Qwen2.5-VL-3B} & Direct & 31.98 & 17.43 & 23.50 & 22.86 & 21.60 & 27.77 & 24.89 & 18.67  \\ 
     & & M3DocRAG & 49.08 & 32.91 & 38.70 & 31.24 & 42.77 & 37.17 & 39.82 & 32.22  \\ 
     &  & MoLoRAG & 50.60 & 36.93 & 38.09 & 32.20 & 44.66 & 39.54 & 41.95 & 34.15  \\ 
     & & MoLoRAG+ &  \textbf{52.06} & \textbf{37.56} & \textbf{40.40} & \textbf{32.97} & \textbf{46.83} & \textbf{39.91} & \textbf{43.17} & \textbf{34.80}  \\ \cmidrule{2-11} 
        
      &  \multirow{4}{*}{Qwen2.5-VL-7B} & Direct & 32.37 & 19.88 & 27.09 & 23.25 & 24.20 & 28.15 & 26.38 & 19.74  \\ 
     & & M3DocRAG & 52.00 & 38.18 & 41.79 & 34.86 & 49.67 & 37.47 & 43.32 & 34.19 \\ 
     & & MoLoRAG & 55.91 & 42.77 & 44.06 & 36.58 & 52.88 & 41.46 & 46.89 & 37.25  \\ 
     & & MoLoRAG+ & \textbf{56.99} & \textbf{42.81} & \textbf{45.77} & \textbf{37.24} & \textbf{53.91} & \textbf{42.47} & \textbf{47.86} & \textbf{37.81} \\ \midrule 

     \textbf{\textit{Multi-agent}} & \multicolumn{2}{c|}{MDocAgent (LLaMA3.1-8B+Qwen2.5-VL-7B)} & 54.52 & 40.41 & 43.20 & 31.25 & 48.88 & 40.26 & 44.42 & 36.30\\
       
      \bottomrule
    \end{tabular}
    
    }
    \label{tab:fine_grained_longdoc}
\end{table*}

\begin{table*}[!t]
    \centering
    \caption{\textbf{Fine-grained performance analysis (Accuracy in $\%$) across evidence modality and evidence locations on LongDocURL under the retrieved top-$3$ setting.}}
    \vspace*{-8pt}
    \resizebox{\linewidth}{!}{

    \begin{tabular}{ccc|cccc|cc|cc}
      \toprule
      \rowcolor{COLOR_MEAN} & &  & \multicolumn{4}{c|}{\textbf{Modality}} & \multicolumn{2}{c|}{\textbf{Location}} & \multicolumn{2}{c}{\textbf{Overall}} \\ 
      \rowcolor{COLOR_MEAN} \multirow{-2}{*}{\textbf{Type}} & \multirow{-2}{*}{\textbf{Model}} & \multirow{-2}{*}{\textbf{Method}} & Text & Table & Figure &  Layout & Single & Multiple  & Acc & EM \\ \midrule
      \multirow{5}{*}{\textbf{\textit{LLM-based}}} & Mistral-7B & Text RAG & 33.94 & 17.95 & 21.25 & 18.98 & 23.09 & 26.63 & 25.06 & 19.78 \\ 
     & Qwen2.5-7B & Text RAG & 36.41 & 20.55 & 25.94 & 23.77 & 26.75 & 28.73 & 27.93 & 21.94  \\ 
     & LLaMA3.1-8B & Text RAG & 37.22 & 22.99 & 29.64 & 22.53 & 29.42 & 30.08 & 29.80 & 22.75 \\ 
     & GPT-4o & Text RAG & 40.91 & 26.80 & 33.14 & 26.57 & 33.19 & 32.13 & 32.74 & 25.20   \\ 
     & DeepSeek-V3 & Text RAG & \textbf{41.89} & \textbf{30.84} & \textbf{35.49} & \textbf{28.15} & \textbf{35.77} & \textbf{33.67} & \textbf{34.73} & \textbf{26.84}  \\  \midrule 

     \multirow{16}{*}{\textbf{\textit{LVLM-based}}} & \multirow{4}{*}{LLaVA-Next-7B} & Direct & 16.79 & 5.28 & 12.12 & 7.39 & 7.87 & 13.43 & 10.78 & 9.29   \\ 
     & & M3DocRAG &  \textbf{20.64} & 7.17 & 16.16 & \textbf{10.75} & 11.12 & 16.20 & \textbf{13.85} & 10.62  \\ 
     & & MoLoRAG & 20.52 & 6.45 & 15.64 & 10.58 & 10.94 & 15.68 & 13.49 & \textbf{10.75} \\ 
     & & MoLoRAG+ & 19.94 & \textbf{7.32} & \textbf{17.11} & 10.64 & \textbf{11.17} & \textbf{15.73} & 13.58 & 10.49  \\ \cmidrule{2-11} 

     & \multirow{4}{*}{DeepSeek-VL-16B} & Direct & 19.98 & 8.26 & 13.81 & 13.65 & 11.18 & 17.87 & 14.72 & 11.35  \\ 
     & & M3DocRAG & 40.61 & 16.19 & 27.56 & 30.78 & 25.54 & 33.31 & 29.60 & 21.29 \\ 
     & & MoLoRAG & 40.62 & 17.57 & 28.69 & 28.86 & 27.07 & 32.67 & 29.98 & 21.81 \\ 
     & & MoLoRAG+ & \textbf{44.28} & \textbf{29.89} & \textbf{37.81} & \textbf{32.84} & \textbf{39.19} & \textbf{35.58} & \textbf{37.21} & \textbf{27.74}  \\ \cmidrule{2-11} 

      &  \multirow{4}{*}{Qwen2.5-VL-3B} & Direct & 31.98 & 17.43 & 23.50 & 22.86 & 21.60 & 27.77 & 24.89 & 18.67  \\ 
      & & M3DocRAG & 54.07 & 37.97 & 42.07 & 36.97 & 46.39 & 42.64 & 44.4 & 34.97   \\ 
      & & MoLoRAG & \textbf{55.99} & 38.23 & \textbf{42.95} & \textbf{37.01} & \textbf{48.09} & \textbf{43.76} & \textbf{45.79} & \textbf{36.17} \\ 
     & & MoLoRAG+ & 54.24 & \textbf{39.03} & 41.13 & 36.62 & 47.49 & 43.31 & 45.27 & 35.53  \\ \cmidrule{2-11}

       & \multirow{4}{*}{Qwen2.5-VL-7B} & Direct & 32.37 & 19.88 & 27.09 & 23.25 & 24.20 & 28.15 & 26.38 & 19.74  \\ 
     & & M3DocRAG & 58.16 & 43.75 & 46.04 & 41.24 & 53.35 & 45.13 & 49.03 & 38.88  \\ 
     & & MoLoRAG & \textbf{61.46} & 45.66 & \textbf{49.06} & \textbf{43.27} & \textbf{55.60} & \textbf{48.30} & 51.71 & \textbf{40.86} \\ 
     & & MoLoRAG+ & 61.43 & \textbf{45.98} & 49.01 & 42.56 & 55.01 & 49.01 & \textbf{51.85} & 40.13  \\ \midrule

     \textit{\textbf{Multi-agent}} & \multicolumn{2}{c|}{MDocAgent (LLaMA3.1-8B+Qwen2.5-VL-7B)} &  56.81 & 42.25 & 44.07 & 35.48 & 49.46 & 44.51 & 46.91 & 37.63 \\
 
      \bottomrule
    \end{tabular}
    
    }
    \label{tab:fine_grained_longdoc_top3}
\end{table*}

\begin{table*}[!t]
    \centering
     \caption{\textbf{Fine-grained performance analysis (Accuracy in $\%$) across evidence modality and evidence locations on LongDocURL under the retrieved top-$5$ setting.}}
     \vspace*{-8pt}
    \resizebox{\linewidth}{!}{

    \begin{tabular}{ccc|cccc|cc|cc}
      \toprule
      \rowcolor{COLOR_MEAN} &  &  & \multicolumn{4}{c|}{\textbf{Modality}} & \multicolumn{2}{c|}{\textbf{Location}} & \multicolumn{2}{c}{\textbf{Overall}} \\ 
      \rowcolor{COLOR_MEAN} \multirow{-2}{*}{\textbf{Type}} & \multirow{-2}{*}{\textbf{Model}} & \multirow{-2}{*}{\textbf{Method}} & Text & Table & Figure &  Layout & Single & Multiple  & Acc & EM \\ \midrule
      \multirow{5}{*}{\textbf{\textit{LLM-based}}} & Mistral-7B & Text RAG & 34.74 & 18.78 & 22.91 & 19.91 & 25.57 & 26.94 & 26.43 & 20.22 \\ 
     & Qwen2.5-7B & Text RAG & 39.38 & 24.63 & 29.76 & 24.72 & 30.48 & 31.92 & 31.36 & 25.03\\ 
     & LLaMA3.1-8B & Text RAG &40.04 & 26.02 & 32.74 & 26.01 & 32.56 & 33.85 & 33.27 & 25.42 \\ 
     & GPT-4o & Text RAG & 44.20 & 31.54 & 40.20 & 29.32 & 37.86 & 36.00 & 36.98 & 28.13  \\ 
     & DeepSeek-V3 & Text RAG & \textbf{45.71} & \textbf{34.39} & \textbf{41.58} & \textbf{31.26} & \textbf{40.09} & \textbf{38.08} & \textbf{39.04} & \textbf{29.38} \\  \midrule 

     \multirow{16}{*}{\textbf{\textit{LVLM-based}}} & \multirow{4}{*}{LLaVA-Next-7B} & Direct & 16.79 & 5.28 & 12.12 & 7.39 & 7.87 & 13.43 & 10.78 & 9.29   \\ 
     & & M3DocRAG & 19.20 & 5.89 & 13.58 & 9.36 & 8.86 & 15.93 & 12.65 & 10.02   \\ 
     & & MoLoRAG & 19.39 & 5.94 & 13.51 & 10.13 & 9.18 & 15.79 & 12.72 & 10.19  \\ 
     & & MoLoRAG+ & \textbf{20.03} & \textbf{7.00} & \textbf{16.67} & \textbf{10.69} & \textbf{10.79} & \textbf{16.01} & \textbf{13.59 }& \textbf{10.58}  \\ \cmidrule{2-11} 

      & \multirow{4}{*}{DeepSeek-VL-16B} & Direct & 19.98 & 8.26 & 13.81 & 13.65 & 11.18 & 17.87 & 14.72 & 11.35  \\ 
     & & M3DocRAG &  40.54 & 15.44 & 26.41 & 30.35 & 24.50 & 33.62 & 29.27 & 21.03 \\ 
     & & MoLoRAG & 42.08 & 16.97 & 28.58 & 31.09 & 26.40 & 34.76 & 30.76 & 22.19 \\ 
     & & MoLoRAG+ & \textbf{46.11} & \textbf{29.48} & \textbf{38.03} & \textbf{34.02} & \textbf{39.76} & \textbf{36.61} & \textbf{38.02} & \textbf{28.34}  \\ \cmidrule{2-11} 

      &  \multirow{4}{*}{Qwen2.5-VL-3B} & Direct & 31.98 & 17.43 & 23.50 & 22.86 & 21.60 & 27.77 & 24.89 & 18.67  \\ 
     & & M3DocRAG &  54.76 & 37.06 & 39.66 & 38.47 & 45.49 & 43.95 & 44.67 & 34.84 \\ 
     & & MoLoRAG & \textbf{55.29} & \textbf{39.68} & 40.79 & \textbf{38.88} & \textbf{47.07} & \textbf{45.25} & \textbf{46.05} & \textbf{35.74} \\ 
     & & MoLoRAG+ &  53.12 & 39.45 & \textbf{40.95} & 37.32 & 47.05 & 43.43 & 45.13 & 35.53 \\ \cmidrule{2-11} 
     
      & \multirow{4}{*}{Qwen2.5-VL-7B} & Direct & 32.37 & 19.88 & 27.09 & 23.25 & 24.20 & 28.15 & 26.38 & 19.74  \\ 
      & & M3DocRAG & 59.52 & 45.01 & 45.25 & 42.82 & 53.14 & 47.71 & 50.33 & 39.23 \\ 
     &  & MoLoRAG & \textbf{60.70} & \textbf{46.99} & 47.95 & 44.74 & \textbf{55.23} & 49.65 & \textbf{52.33} & \textbf{41.76} \\ 
     &  & MoLoRAG+ & 60.54 & 46.61 & \textbf{48.68} & \textbf{45.13} & 54.86 & \textbf{50.05} & 52.33 & 40.65  \\ \midrule 

     \textbf{\textit{Multi-agent}} & \multicolumn{2}{c|}{MDocAgent (LLaMA3.1-8B+Qwen2.5-VL-7B)}  &57.09 & 44.66 & 45.97 & 35.47 & 50.79 & 45.52 & 48.07 & 38.32 \\ 

      \bottomrule
    \end{tabular}
    
    }
    \label{tab:fine_grained_longdoc_top5}
\end{table*}